\PassOptionsToPackage{dvipsnames}{xcolor}

\documentclass[10pt,twocolumn,letterpaper]{article}
\pdfoutput=1

\usepackage{cuted}
\usepackage{dblfloatfix} 
\usepackage{afterpage} 
\usepackage{caption}
\usepackage{float} 
\usepackage{algorithm2e}
\RestyleAlgo{ruled}
\usepackage{placeins}
\usepackage{multirow,bigdelim}
\usepackage{booktabs} % For formal tables
\usepackage{makecell}
\usepackage{tikz}
\usepackage{pifont}
\usepackage{svg}
\usepackage{amsmath}

\newcommand{\myparagraph}[1]{\noindent\textbf{#1}}
\usepackage[pagenumbers]{cvpr} 

\definecolor{cvprblue}{rgb}{0.21,0.49,0.74}
\usepackage[pagebackref,breaklinks,colorlinks,allcolors=cvprblue]{hyperref}

\newcommand{\method}{\textsc{Through-The-Mask}\xspace}
\newcommand{\benchmark}{\textsc{SA-V-128}\xspace}

\title{Through-The-Mask: Mask-based Motion Trajectories for \\ Image-to-Video Generation}

\author{
Guy Yariv$^{1,3}$ \quad Yuval Kirstain$^{1}$ \quad Amit Zohar$^{1}$ \quad Shelly Sheynin$^{1}$ \quad Yaniv Taigman$^1$ \\
Yossi Adi$^{2,3}$ \quad Sagie Benaim$^{3}$ \quad Adam Polyak$^1$ \\
\\
$^1$GenAI, Meta \quad $^2$FAIR, Meta \quad $^3$The Hebrew University of Jerusalem \\
}

\begin{document}

\twocolumn[{%
\renewcommand\twocolumn[1][]{#1}%
\maketitle

\begin{center}
    \vspace{-2.0em}
    \centering
    \captionsetup{type=figure}
    \includegraphics[width=1.0\linewidth]{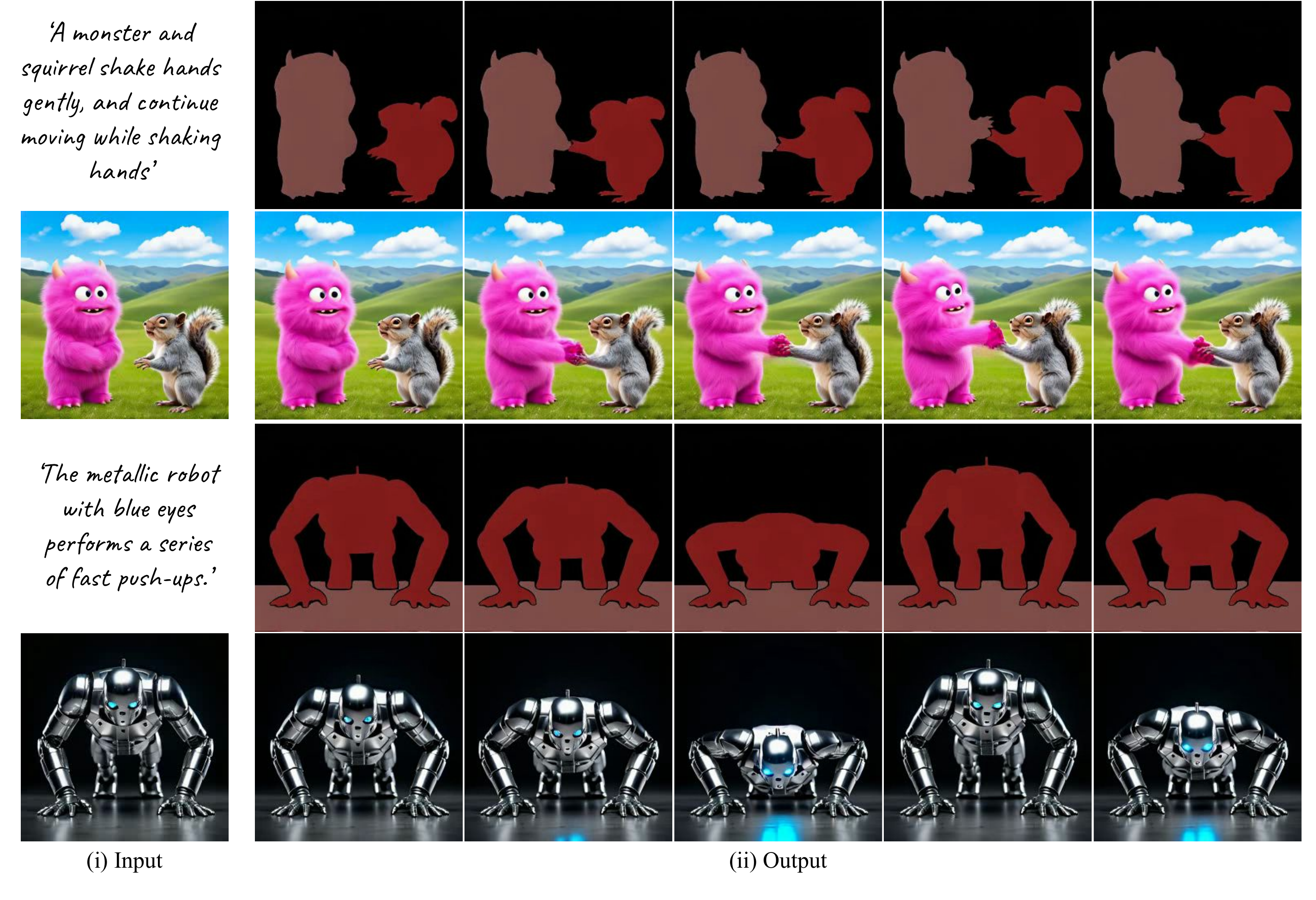}
    \vspace{-2.0em}
    \captionof{figure}{
    \method is an Image-to-Video method that animates an input image based on a provided text caption. The generated video (rows 2 and 4) leverages mask-based motion trajectories (rows 1 and 3), enabling accurate animation of multiple objects. }
    \label{fig:teaser}
\end{center}

}]

\begin{abstract}
We consider the task of Image-to-Video (I2V) generation, which involves transforming static images into realistic video sequences based on a textual description.
While recent advancements produce photorealistic outputs, they frequently struggle to create videos with accurate and consistent object motion, especially in multi-object scenarios.
To address these limitations, we propose a two-stage compositional framework that decomposes I2V generation into: (i) An explicit intermediate representation generation stage, followed by (ii) A video generation stage that is conditioned on this representation. 
Our key innovation is the introduction of a mask-based motion trajectory as an intermediate representation, that captures both semantic object information and motion, enabling an expressive but compact representation of motion and semantics. To incorporate the learned representation in the second stage, we utilize object-level attention objectives. Specifically, we consider a spatial, per-object, masked-cross attention objective, integrating object-specific prompts into corresponding latent space regions and a masked spatio-temporal self-attention objective, ensuring frame-to-frame consistency for each object. We evaluate our method on challenging benchmarks with multi-object and high-motion scenarios and empirically demonstrate that the proposed method achieves state-of-the-art results in temporal coherence, motion realism, and text-prompt faithfulness. Additionally, we introduce \benchmark, a new challenging benchmark for single-object and multi-object I2V generation, and demonstrate our method's superiority on this benchmark. Project page is available at \href{https://guyyariv.github.io/TTM/}{https://guyyariv.github.io/TTM/}.

\end{abstract}    
\section{Introduction}
\label{sec:intro}

Image-to-Video~(I2V) generation transforms static images into realistic video sequences guided by textual descriptions.
Recently, significant progress has been made in this task, with models such as \cite{dai2023animateanythingfinegrainedopendomain, blattmann2023stablevideodiffusionscaling, ma2024cinemoconsistentcontrollableimage, ren2024consisti2venhancingvisualconsistency, shi2024motioni2vconsistentcontrollableimagetovideo, zhang2023i2vgenxlhighqualityimagetovideosynthesis}, which enable the generation of photorealistic and consistent output. However, current works still struggle to generate videos with consistent and faithful object motion. 
These limitations are especially evident in scenarios with multiple objects, as shown in our experiments, where capturing the correct motion and interactions is challenging.

Several works~\cite{blattmann2023stablevideodiffusionscaling, ma2024cinemoconsistentcontrollableimage, dai2023animateanythingfinegrainedopendomain, ren2024consisti2venhancingvisualconsistency} directly map an input image (and possibly text) to an output video in a single, end-to-end pipeline. By doing so, the underlying model must implicitly reason about object semantics and motion while simultaneously generating a plausible appearance of all objects.  
As the range of possible motions and interactions scales significantly with the number of objects, this makes it difficult for current models to generate plausible outputs. 
An alternative approach is to decompose the training process into a two-stage compositional process: (i) Given the input image, generate an \textit{explicit} intermediate representation; (ii) Utilize the generated representation and the input image to generate the full video. Recent work~\citep{shi2024motioni2vconsistentcontrollableimagetovideo}, proposed using Optical Flow (OF) for this representation. However, this has several drawbacks. First, only motion, without semantics, is represented in OF. Second, such motion representation is redundant in the context of I2V generation. Predicting per-point pixel motion may not be required to depict plausible object motion. Doing so may result in unnecessary errors (e.g., wrong prediction in pixels from non-moving objects). Such errors can then significantly influence the second stage, as the model tries to predict the correct appearance while adhering to incorrect motion. 

In this work, we argue that the choice of representation is critical and should capture several properties: (i) it should express both motion and semantics; (ii) it should represent the motion and interaction of individual objects; and (iii) it should be robust to signal variations and operate at the object level rather than at the pixel level. We argue that a suitable choice satisfying these properties is a \textit{mask-based motion trajectory}, a time-consistent per-frame semantic mask capturing semantic objects and their motion~(see Fig.~\ref{fig:teaser}). 

Our method follows a two-stage process: first, a network is trained to generate a mask-based motion trajectory conditioned on the input image, segmentation mask, and text prompt. In the second stage, the motion-to-video network generates the final video conditioned on the input image, text prompt, and the generated motion trajectory from the first stage.

For the second stage of our framework (i.e., motion-to-video), we propose to integrate the generated \textit{mask-based motion trajectory}'s structure 
softly, using learned attention layers, ensuring the network adheres to the generated semantics and motion.  Specifically, we propose using (i) a \textit{masked cross-attention} objective, which integrates object-specific prompts directly into corresponding regions of the latent space, using masked cross-attention, and (ii) a \textit{masked self-attention} objective, which ensures that each object maintains consistency across frames, by using the generated mask in the self-attention mechanism to restrict attention to positions corresponding to the same object. 

We compare our approach to a diverse set of recent image-to-video generation approaches on challenging benchmarks that include several objects and significant motion. We demonstrate state-of-the-art performance across diverse metrics, including temporal coherence, motion realism, visual consistency across frames, and text faithfulness to the input prompt. To further advance research on I2V generation, we introduce a new benchmark that includes distinct sets for single-object and multi-object videos, demonstrating superior performance. Finally, we ablate our method, demonstrating the contribution of each component. 
\section{Related work}
\label{sec:related}

\myparagraph{Text-to-Video Generation.} Recent advances in diffusion models~\cite{ho2020denoising,song2020denoising,song2020score} and flow matching techniques~\cite{lipman2023flowmatchinggenerativemodeling,albergo2023stochastic,liu2022rectified} have enhanced the capability to generate high-quality images conditioned on textual descriptions~\cite{dalle3,emu,stable3}. In the context of text-conditioned video generation, several approaches perform diffusion in a low-dimensional latent space, adopting the Latent Diffusion Models~(LDM) architecture~\cite{rombach2022high,podell2023sdxl}. Many text-to-video~(T2V) models adapt T2I architectures to generate temporally coherent videos, extending beyond the spatial knowledge of T2I training~\cite{ho2022video,singer2022makeavideotexttovideogenerationtextvideo,ho2022imagen,ge2023preserve,wang2023lavie,blattmann2023align,an2023latent,wang2023videofactory,wang2023modelscope,zhou2022magicvideo,he2022latent,chen2024videocrafter2,guo2024animatediffanimatepersonalizedtexttoimage}. A common approach extends pre-trained T2I models with temporal modules, such as convolutions or attention layers, followed by additional training for video generation~\cite{ho2022video,singer2022makeavideotexttovideogenerationtextvideo,ho2022imagen,ge2023preserve}. EmuVideo~\cite{girdhar2023emu} and VideoGen~\cite{li2023videogen}, for instance, factorized the text-to-video to two stages: text-to-image and image-to-video. 
Recent studies have adopted the transformer-based Diffusion Transformer~(DiT)~\cite{peebles2023scalable} due to its performance over U-Net~\cite{ma2024latte,menapace2024snap,sora,polyak2024moviegencastmedia}. Notably, our approach is architecture-agnostic and works with both U-Net and DiT. %backbones.

\myparagraph{Image-To-Video Generation.} In I2V, the video generation model is conditioned on the input text, as well as an additional visual input that represents the initial frame of the output video~\cite{blattmann2021understanding, pan2019video}. Several works leverage this additional visual input by fine-tuning of a pre-trained T2V model~\cite{chen2023videocrafter1opendiffusionmodels,blattmann2023stablevideodiffusionscaling}. Despite encouraging progress in generated video aesthetic~\cite{girdhar2023emu}, current I2V models struggle to generate complex actions or interactions between objects~\cite{shi2024motioni2vconsistentcontrollableimagetovideo}.

Recent work has attempted to tackle this challenge. VideoCrafter~\cite{chen2023videocrafter1opendiffusionmodels} incorporate an additional image input to preserve the content and style of the reference image. DynamiCrafter~\cite{xing2023dynamicrafteranimatingopendomainimages} use a query transformer to project the image into a text-aligned context, leveraging motion priors from T2V models to animate images. I2VGen-XL~\cite{zhang2023i2vgenxlhighqualityimagetovideosynthesis} employ a two-stage cascade, with the second stage refining resolution and temporal coherence. ConsistI2V~\cite{ren2024consisti2venhancingvisualconsistency} use first-frame conditioning by combining the initial latent frame with input noise and enhancing self-attention with intermediate states. AnimateAnything~\cite{dai2023animateanythingfinegrainedopendomain} include an additional mask to constrain motion areas. Cinemo~\cite{ma2024cinemoconsistentcontrollableimage} introduce three enhancements: (i) prediction of motion residuals, (ii) fine-grained control over motion intensity, and (iii) noise refinement at inference to reduce motion shifts. Our work considers a different two-step compositional approach and is orthogonal to these advances.

Perhaps most relevant to our work is Motion-I2V~\cite{shi2024motioni2vconsistentcontrollableimagetovideo}, which follows a two-step generation process: (i) prediction of optical flow displacement, and (ii) generation of the video based on generated optical flow. Our method similarly follows a two-stage approach. However, we offer two key differences. First, we use a different intermediate representation of mask-based motion trajectories. We argue that this choice is significant in multiple aspects: (i) we represent not only motion, but also semantics, enhancing expressivity. (ii) Simultaneously, our representation captures only object-level motion as opposed to pixel-level motion. Doing so makes our generation less susceptible to errors in the first stage, as also observed in~\cite{avrahami2023spatext}. Second, our representation also enables additional flexibility in the second generation stage, which generates a video conditioned on this representation. Specifically, instead of wrapping the generated video using predicted flow, we softly condition the video generation model on the intermediate representation using object-level and temporal attention. 
We note that recent T2I models have introduced conditioning on specific areas with targeted information to improve fine-grained controllability~\citep{nie2024compositionaltexttoimagegenerationdense, li2023gligenopensetgroundedtexttoimage}. To address the I2V setting, we extend the masked cross-attention introduced in \cite{nie2024compositionaltexttoimagegenerationdense} to the video setting and introduce a novel masked self-attention objective.

\begin{figure*}[t!]
    \centering
    \includegraphics[width=0.96\textwidth]{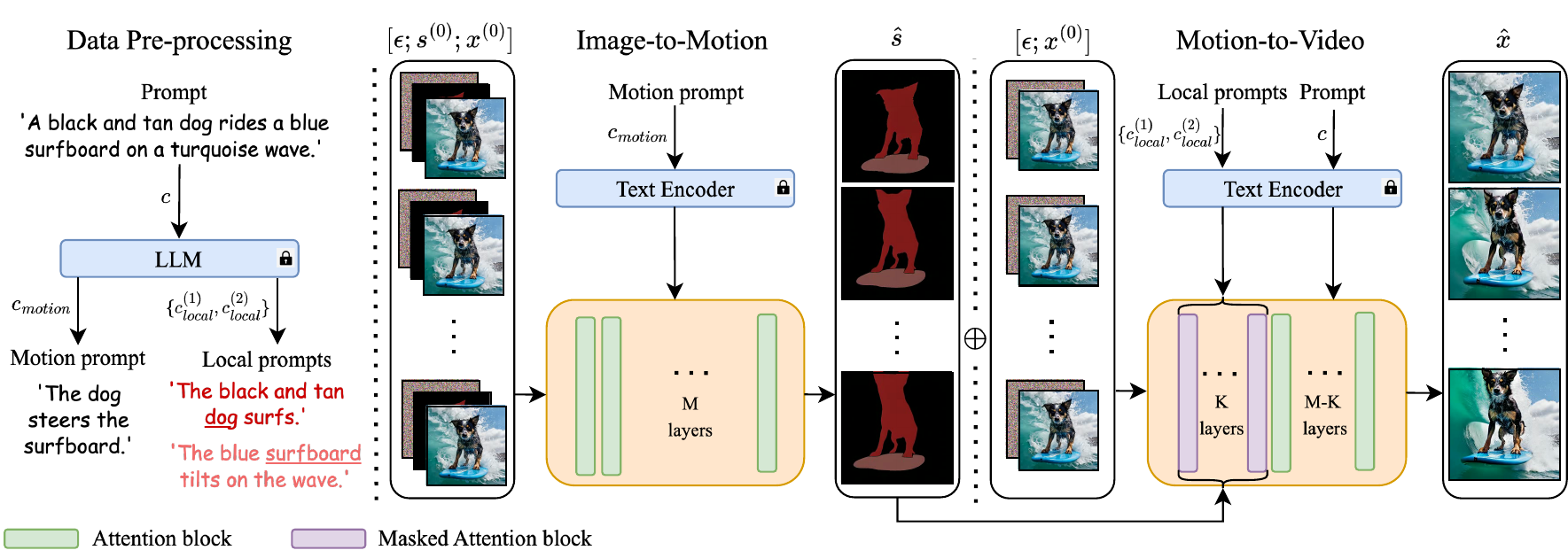}
    \caption{
    Overview of our I2V framework, transforming a reference image \( x^{(0)} \) and text prompt \( c \) into a coherent video sequence \( \hat{x} \). A pre-trained LLM is used to derive the motion-specific prompt \( c_{motion} \) and object-specific prompts \( c_{local} = \{c_{local}^{(1)}, \dots, c_{local}^{(L)}\} \), capturing each object's intended motion. We generate an initial segmentation mask \( s^{(0)} \) from \( x^{(0)} \) using SAM2. In Stage 1, the Image-to-Motion utilizes \( x^{(0)} \), \( s^{(0)} \), and \( c_{motion} \) to generate mask-based motion trajectories \( \hat{s} \) that represent object-specific movement paths. In Stage 2, the Motion-to-Video takes as input \( x^{(0)} \), the generated trajectories \( \hat{s} \), the text prompt \( c \) as a global condition, and object-specific prompts \( c_{local} \) through a masked attention blocks (Section~\ref{sec:motion_to_video}), producing the final video \( \hat{x} \).
    }
    \label{fig:method}
    \vspace{-0.5cm}
\end{figure*}

\section{Method}

Our method \method, illustrated in Fig.~\ref{fig:method}, 
factorizes I2V into two compositional stages:
\begin{enumerate}
    \item \textbf{Image-to-Motion Generation:} In the first stage, outlined in Sec.~\ref{sec:image_to_motion}, we generate motion trajectory conditioned on the reference image and motion prompt. This motion trajectory encapsulates the dynamic behavior of individual objects.
    \item \textbf{Motion-to-Video Generation:} In the second stage, outlined in Sec.~\ref{sec:motion_to_video}, we use the generated motion trajectory, along with the object-specific prompts and the reference image, to produce a photorealistic video.
\end{enumerate}
Our two-stage process is based on the choice of an explicit intermediate representation of objects' motion. Ideally, such representation should (i) express both motion and semantics, (ii) represent the interaction of objects, and (iii) be robust to signal variations. We claim that a motion trajectory, i.e., a \emph{consistent per-frame video segmentation}, satisfies these properties by definition. First, it captures motion, interactions, and type (i.e., semantics), hence satisfying both (i) and (ii). 
Second, as image segmentation operates at a coarse level (i.e., object level rather than pixel level), we achieve the following desired separation of tasks: The first stage handles coarse object-level semantics, motion, and inter-object interactions. The second stage then handles the intra-object level semantics and appearance. On average, this results in fewer overall errors produced than the pixel-level counterpart in the first stage, satisfying (iii). It also enables greater flexibility in the second stage.

To allow the modeling of our framework, we pre-process our training data, as outlined in  
Sec.~\ref{sec:motion_trajectories}.
The Image-to-Motion and Motion-to-Video stages are trained independently but are combined during inference to produce the final video (see supplementary Sec.~\ref{sec:inference}).
Additional implementation details are provided in supplementary Sec.~\ref{sec:implementation}.

\subsection{Data Pre-processing} \label{sec:motion_trajectories}

We assume a training dataset of text-video pairs, where the input contains a reference image $x^{(0)}$ and a text prompt $c$. Our pre-processing pipeline comprises the following components applied on each text-video pair: (i) extraction of prompts for motion-capable objects from the input text, (ii) video segmentation, and (iii) extraction of motion-specific and object-specific prompts from the input text. %%We describe each of them below.

\myparagraph{Motion-capable Object Prompt Extraction.} Using a pre-trained Large Language Model (LLM), we extract $L$ objects' prompts, $\{o^{(1)}, \dots, o^{(L)} \}$, relevant to generating specific motion pathways (motion-capable objects) from the input text, $c$. Notice, $L$ is variable and video-specific. We refer the readers to supplementary Sec.~\ref{sec:object_capable} for more details.
\myparagraph{Video Segmentation.} 
We assume an $N$-frame video, $x=\{x^{(0)}, \dots, x^{(N)}\}$, where $x^{(i)} \in \mathbb{R}^{3 \times H \times W}$ are  frames of resolution $H \times W$. 
To obtain a trajectory, we first use Grounding DINO~\cite{liu2024groundingdinomarryingdino} to obtain bounding boxes for each motion-capable object caption $o^{(i)}$ within the first frame, $x^{(0)}$.
Using these bounding boxes as input, we use SAM2~\cite{ravi2024sam2segmentimages} to create video segmentation $s=\{s^{(0)}, \dots, s^{(N)}\}$, where mask $s^{(i)}$ matches video frame $x^{(i)}$.

\myparagraph{Motion and Object-Specific Prompts.} Using an LLM, we extract two variants of text prompts to guide motion generation: (i) a motion-specific prompt $c_{motion}$ that consolidates all motion information without spatial details, and (ii) a set of object-specific prompts $c_{local}=\{c_{local}^{(1)}, \dots, c_{local}^{(L)}\}$, where each prompt provides details specific to each object’s motion (see Fig.~\ref{fig:method} for an example).
By assigning each object a constant color in the mask trajectory, we can reliably match each object-specific prompt to its spatial location at any time.  
Additional prompt generation details are provided in supplementary Sec.~\ref{sec:motion_and_object_prompt}.
Following pre-processing, each data sample consists of the tuple $(x, s, c, c_{motion}, c_{local})$, where $x$ is the ground truth video, $s$ the segmentation, $c$ the initial text prompt, $c_{motion}$ the motion-specific prompt, and $c_{local}$ the set of object-specific prompts.

\subsection{Image-to-Motion} \label{sec:image_to_motion}
In the first stage of our framework, we train a model to generate a sequence of fine-grained, mask-based motion trajectories, $\hat{s}$, conditioned on an input frame $x^{(0)}$, a segmentation of the input frame $s^{(0)}$, and a motion-specific prompt $c_{motion}$.
The Image-to-Motion model is denoted as $\hat{s}_{\theta}(s_t, t, \mathcal{E}(x^{(0)}),\mathcal{E}(s^{(0)}),c_{motion})$, where $s_t$ is a noisy masked-based motion trajectory at the denoising timestep $t$, $\theta$ are the learned parameters of the network, and $\mathcal{E}$ is a VAE~\cite{kingma2013auto} encoder.
For brevity, we omit the activation of the encoder, $\mathcal{E}$, in the rest of the section. We apply a denoising process in the latent space of a VAE as in LDM~\cite{rombach2022high}. We initialize $\hat{s}_{\theta}$ with a pre-trained text-to-video model by concatenating the encodings of the first frame $x^{(0)}$ and its mask $s^{(0)}$ along the input channel dimension. 
Text conditioning is applied as in LDM, using cross-attention layers. 
See the supplementary for full details. 

\subsection{Motion-to-Video} \label{sec:motion_to_video}
In the second stage, we train a model to generate a sequence of video frames $\hat{x}$, conditioned on the reference image $x^{(0)}$, the generated motion trajectory $\hat{s}$, the text prompt $c$, and the object-specific prompts $c_{local}$. 
Using the same denoising approach as in the first stage, we train the Motion-to-Video model, $\hat{x}_\psi(x_{t}, t, x^{(0)},\hat{s},c,c_{local})$ with parameters $\psi$, to iteratively refine a noisy latent representation towards a clean video output, following LDM~\cite{rombach2022high}'s formulation. 
As in the first stage, we finetune a pre-trained text-to-video model, concatenating the encodings of the first frame $x^{(0)}$ and the predicted mask-based motion trajectory $\hat{s}$ to the noisy latent representation $x_{t}$ along the channel dimension. Text is integrated using cross-attention as in the first stage. 

\begin{figure}[t!]
    \centering
    \includegraphics[width=0.4\textwidth]{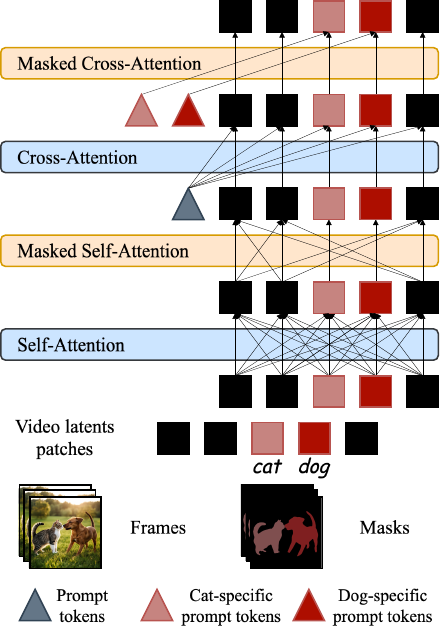}
    \caption{\textbf{Illustration of the masked attention block.} Squares represent video latent patches, color-coded to indicate objects (e.g., cat or dog). Triangles denote prompt tokens: gray for global prompts and object-specific colors for local prompts. The pipeline features self-attention for all patches, masked self-attention restricted to each object, cross-attention integrating global prompts, and masked cross-attention aligning object-specific prompts.}
    \label{fig:masked_attn}
    \vspace{-0.4cm}
\end{figure}

\subsubsection{Masked Attention Blocks} 
We introduce two masked attention-based objectives to condition I2V models in specific areas with targeted information, as shown in Fig.~\ref{fig:masked_attn}. 
We apply these objectives in the first $K$ blocks to extend the model’s attention capabilities. 
In the following section, we build upon the notation of~\cite{nie2024compositionaltexttoimagegenerationdense}.

\myparagraph{Masked Cross-Attention.} We employ masked cross-attention to integrate object-specific prompts directly into the corresponding regions of the latent space, ensuring each object's latent representation attends only to its own prompt. Our approach extends that of \cite{nie2024compositionaltexttoimagegenerationdense}, which considered an object-level cross-attention for text-to-image generation. 

Formally, let $z \in \mathbb{R}^{N' \times H' \times W' \times d}$ be the latent features serving as queries, where $N',H',W'$ are the temporal and spatial dimension of the latent features and $d$ is the model dimension size.
For $L$ object-specific prompts $\{c_{local}^{(i)}\}_{i=1}^L$, we encode the prompts to obtain a sequence of embeddings $\{e^{(i)}\}_{i=1}^L \in \mathbb{R}^{N_{txt} \times d}$, where $N_{txt}$ is the sequence length of the encoded prompts.
We denote the query, key, and value of the masked-cross attention layers as follows $q = z W_q \in \mathbb{R}^{N_{tokens} \times d}$, $k^{(i)} = e^{(i)} W_k \in \mathbb{R}^{N_{txt} \times d}$, and $v^{(i)} = e^{(i)} W_v \in \mathbb{R}^{N_{txt} \times d}$, where $N_{tokens}=N' \cdot H'\cdot W'$. 
All object-specific keys and values are concatenated along the sequence dimension, $k = [k^{(1)}; \dots; k^{(L)}]$ and $v = [v^{(1)}; \dots; v^{(L)}]$. 
The masked cross-attention then becomes,
\begin{align}
% k &= [k^{(1)}; \dots; k^{(L)}] \\
% v &= [v^{(1)}; \dots; v^{(L)}] \\
M_{cross} &= [M^{(1)}; \dots; M^{(l)}] \\
h_{cross}&= \sigma \left( \frac{qk^T}{\sqrt{d}} + \log{M_{cross}} \right)v,
\end{align}
where $[\cdot; \cdot]$ is a concatenation along the sequence dimension, $\sigma(\cdot)$ is the softmax function, and $h$ is the intermediary hidden features passed to the next layer.
We construct binary masks \( M^{(l)} \in \{0,1\}^{N_{tokens}} \) indicating the spatial locations associated with each object \( l \) along the frames, derived from bounding boxes obtained during training (from ground truth segmentation \( s \)) or inference (from generated segmentation \( \hat{s} \)). The masked cross-attention is computed by restricting each query position to attend only to the keys corresponding to objects present at that location. 

\myparagraph{Masked Self-Attention.}
Unlike cross-attention, where the queries come from one sequence and the keys and values come from another, self-attention derives the query $q$, key $k$, and value $v$ from the same input sequence, which is the latent features $z$. 
We introduce a novel objective
that ensures that 
each position attends only to positions of the same object, enhancing temporal consistency and preventing interference between different objects.
To this end, we introduce a mask into the self-attention mechanism that restricts attention to positions corresponding to the same object. 
We construct an attention mask \( M_{self} \in \{0,1\}^{N_{tokens} \times N_{tokens}} \), where \( M_{self}^{(i,j)} = 1 \) if positions \( i \) and \( j \) belong to the same object (based on the segmentation masks \( \hat{s} \)), and \( M_{self}^{(i,j)} = 0 \) otherwise. 
The masked self-attention then becomes,
\begin{align}
h_{self} &= \sigma \left( \frac{qk^T}{\sqrt{d}} + \log{M_{self}} \right)v.
\end{align}

Applying this attention mask we get masked self-attention. 

\setlength{\tabcolsep}{2pt}
\begin{table*}[h!]
\centering
\scalebox{0.72}{
\centering
\begin{tabular}{lc@{\hspace{10pt}}cccc|ccc|c@{\hspace{10pt}}cccc|ccc}
\toprule
 & \multicolumn{8}{c|}{Single-Object} & \multicolumn{8}{c}{Multi-Object} \\
\midrule
Method & $\text{FVD}\downarrow$ &  $\text{CF}\uparrow$ & $\text{ViCLIP-T}\uparrow$ & $\text{ViCLIP-V}\uparrow$ & AD & Text & Motion & Quality & $\text{FVD}\downarrow$ &  $\text{CF}\uparrow$ & $\text{ViCLIP-T}\uparrow$ &  $\text{ViCLIP-V}\uparrow$ & AD & Text & Motion & Quality \\
& & & & & & align. & consist. & & & & & & & align. & consist. & \\
\midrule
VideoCrafter~\citep{chen2024videocrafter2} & 1484.18 & 0.966 & 0.209 & 0.796 & 2.93 &  84.3 & 84.3 & 81.2 & 1413.83 & 0.966 & 0.208 & 0.802 & 3.75 & 84.3 & 87.5 & 92.1 \\
DynamiCrafter~\citep{xing2023dynamicrafteranimatingopendomainimages} & 1442.48 & 0.942 & 0.214 & 0.817 & 8.94 & 75.0 & 81.2 & 82.8 & 1300.07 & 0.947 & 0.211 & 0.834 & 7.56 & 75.0 & 73.4 & 76.5 \\
Motion-I2V~\citep{shi2024motioni2vconsistentcontrollableimagetovideo} & 1195.08 & 0.937 & 0.220 & 0.822 & 6.28 & 75.0 & 89.0 & 93.7 & 1162.06 & 0.935 & 0.219 & 0.821 & 6.97 & 81.0 & 89.0 & 95.3 \\
ConsistI2V~\citep{ren2024consisti2venhancingvisualconsistency} & 1206.61 & 0.951 & 0.218 & 0.839 & 5.21 & 65.6 & 78.1 & 81.2 & 1186.10 & 0.935 & 0.217 & 0.850 & 7.25 & 81.2 & 82.8 & 84.3 \\
TI2V~(UNet) & 1285.99 & 0.942 & 0.219 & 0.877 & 5.90 & 53.1 & 59.3 & 70.3 & 1410.68 & 0.942 & 0.218 & 0.883 & 7.93 & 62.5 & 64.0 & 60.0\\
Ours~(UNet) & \textbf{925.39} & \textbf{0.969} & \textbf{0.220} & \textbf{0.888} & 4.70 & - & - & - & \textbf{1089.86} & \textbf{0.966} & \textbf{0.220} & \textbf{0.896} & 5.59 & - & - & - \\
\midrule
TI2V~(DiT) & 1232.89 & 0.924 & 0.223 & 0.797 & 10.87 & 65.6 & 73.4 & 68.7 & 1156.82 & 0.917 & 0.221 & 0.805 & 10.52 & 64.0 & 59.3 & 64.0 \\
Ours~(DiT) & \textbf{1216.83} & \textbf{0.945} & \textbf{0.226} & \textbf{0.860} & 7.22 & - & - & - & \textbf{1134.71} & \textbf{0.948} & \textbf{0.225} & \textbf{0.863} & 7.48 & - & - & - \\
\bottomrule
\end{tabular}}
\caption{Results for single-object and multi-object settings on the \textbf{\benchmark Benchmark}. We report FVD, CLIPFrame~(CF), ViCLIP-T, ViCLIP-V, and Average Displacement~(AD), along with human ratings. Human evaluation shows the percentage of raters that prefer the results of \method.
}
\vspace{-0.5cm}
\label{tab:sav_benchmark}
\end{table*}

\section{Experiments}
To evaluate our method, we assess temporal coherence, motion realism, visual consistency across frames, and text faithfulness. 
First, we compare our approach with current state-of-the-art image-to-video methods on the Image-Animation-Bench, featuring 2,500 high-quality videos.
We use two different neural network architectures for the denoising network: a U-Net, adapted from AnimateDiff~\cite{guo2024animatediffanimatepersonalizedtexttoimage}, and a DiT, adapted from Movie~Gen~\cite{polyak2024moviegencastmedia}.
Following this, we ablate our method's design. 
Image-to-video examples are presented in Fig.~\ref{fig:comparison} with additional samples and qualitative comparisons in the supplementary.
We additionally introduce a new benchmark, \benchmark, for image-to-video generation, which includes distinct sets for single-object and multi-object videos. This separation enables focused testing on both scenarios.

\subsection{Experimental Setup}
\myparagraph{Evaluation Benchmarks.}
\label{sec:benchmarks}
To evaluate the effectiveness of our method, we introduce the \benchmark benchmark, designed to test performance across both single- and multi-object animations in diverse scenarios. 
Current image-to-video benchmarks lack explicit distinctions between single- and multi-object cases, particularly when assessing “motion-capable objects” such as humans and animals. 
This limitation hinders accurate evaluation of models in complex multi-object animations.
To address this, we constructed a balanced test set of 128 videos from the SA-V dataset~\cite{ravi2024sam2segmentimages}, with equal representation of single-object and multi-object cases (64 videos each), averaging 14 seconds per video.
The filtering process consisted of generating text captions and categorizing each video from a set of predefined categories using Llama v3.2-11B~\citep{dubey2024llama3herdmodels}, based on selected frames. 
Each video was then assigned aesthetic and motion scores, with motion quantified by optical flow magnitude via RAFT~\cite{teed2020raftrecurrentallpairsfield}. 
From this, the 500 videos with the highest combined scores were automatically selected, and 64 single-object and 64 multi-object videos were randomly chosen from this set. 
We provide further details in the supplementary Sec.~\ref{sec:benchmark_construction}.
Additionally, we use the Image-Animation-Bench, a curated collection of 2,500 videos, all meeting strict resolution requirements and filtered based on aesthetic standards. Further details are provided in the supplementary Sec.~\ref{sec:i2v_bench}.
We then evaluate our method’s effectiveness across diverse scenarios using both the Image-Animation-Bench and the \benchmark benchmark.

\myparagraph{Evaluation Metrics.}
The objective of image-to-video generation is to produce videos that are high-quality, temporally consistent, faithful to the input text, and maintain key elements of the initial input image across frames. We assess these aspects using both objective and subjective metrics.
For video realism, we employ Fréchet Video Distance~(FVD)~\cite{unterthiner2019accurategenerativemodelsvideo}, which measures the visual disparity between feature embeddings of generated and reference videos. 
To evaluate temporal consistency, we use CLIPFrame (Frame Consistency)~\citep{wu2023cvpr2023textguided}, which computes the average cosine similarity between CLIP~\citep{radford2021learningtransferablevisualmodels} embeddings of individual frames to measure frame-to-frame coherence.
To verify text faithfulness, we use ViCLIP-T, a metric based on ViCLIP~\citep{wang2024internvidlargescalevideotextdataset}, a video CLIP model that incorporates temporal information when processing videos. 
ViCLIP-T calculates the cosine similarity between text and video embeddings, measuring how well the generated video aligns with the input text prompt.
For image faithfulness, we use ViCLIP-V, which, similar to ViCLIP-T, measures cosine similarity between the ViCLIP embeddings of the generated video and a reference video derived from the input image to ensure that the generated video maintains essential visual elements of the initial input image across frames. Given the generation setting of a maximum of 128 frames, this metric ensures relative alignment with the reference video, supporting fidelity to the original input.
Furthermore, we report the Average Displacement by taking the average magnitude of the OF vector between consecutive frames to estimate the degree of dynamics. In this metric, we ensure that the videos exhibit realistic motion by maintaining displacement levels that are neither excessively high nor unnaturally low.
For subjective evaluation, we rely on human raters to compare our approach against the baselines. We present the raters with the input frame, a caption describing the motion, and two generated videos: one from our method and one from the baseline.
The raters are tasked with answering three questions: (i) Text faithfulness: \emph{Which video better matches the caption?} (ii) Motion: \emph{Which video has the best overall motion consistency?} and (iii) Quality: \emph{Aesthetically, which video is better?}
To ensure a fair comparison, each pair of videos, along with the input image and text prompt, was rated by 5 different raters. 
We rate 128 randomly sampled samples from the Image-Animation-Bench and all 128 samples from \benchmark benchmark.
We then used the majority vote to determine which video was preferred.
\begin{table}[t]
\centering
\scalebox{0.65}{
\centering
\begin{tabular}{lccccc|ccc}
\toprule
Method & $\text{FVD}\downarrow$ &  $\text{CF}\uparrow$ & $\text{ViCLIP-T}\uparrow$ &  $\text{ViCLIP-V}\uparrow$ & AD & Text & Motion & Quality  \\
& & & & & & align. & consist. & \\
\midrule
VideoCrafter~\cite{chen2024videocrafter2} & 266.83 & 0.961 & 0.195 & 0.810 & 4.87 & 78.9 & 78.1 & 80.4 \\
DynamiCrafter~\cite{xing2023dynamicrafteranimatingopendomainimages} & 217.40 & 0.946 & 0.200 & 0.840 & 8.28 & 55.4 & 53.9 & 53.9 \\
Motion-I2V~\cite{shi2024motioni2vconsistentcontrollableimagetovideo} & 286.42 & 0.928 & 0.209 & 0.746 & 7.46 & 81.2 & 82.8 & 83.5 \\
ConsistI2V~\cite{ren2024consisti2venhancingvisualconsistency} & 283.59 & 0.938 & 0.202 & 0.838 & 6.38 & 57.0 & 67.1 & 65.6 \\
TI2V~(UNet) & 242.18 & 0.954 & 0.203 & 0.858 & 5.99 & 49.2 & 57.8 & 66.4 \\
Ours~(UNet) & \textbf{196.23} & \textbf{0.962} & \textbf{0.210} & \textbf{0.865} & 5.69 & - & - & - \\
\midrule
TI2V~(DiT) & 212.23 & 0.937 & 0.206 & 0.789 & 9.00 & 61.7 & 63.2 & 67.1 \\
Ours~(DiT) & \textbf{192.45} & \textbf{0.948} & \textbf{0.215} & \textbf{0.847} & 7.42 & - & - & - \\
\bottomrule
\end{tabular}}
\caption{\textbf{Image-Animation-Bench} results. We report FVD, CLIPFrame~(CF), ViCLIP-T, ViCLIP-V, and Average Displacement~(AD), along with human ratings. Human evaluation shows the percentage of raters that prefer the results of \method.}
\vspace{-0.5cm}
\label{tab:i2v_bench}
\end{table}

\begin{figure*}[ht!]
    \centering
    \includegraphics[width=0.97\textwidth]{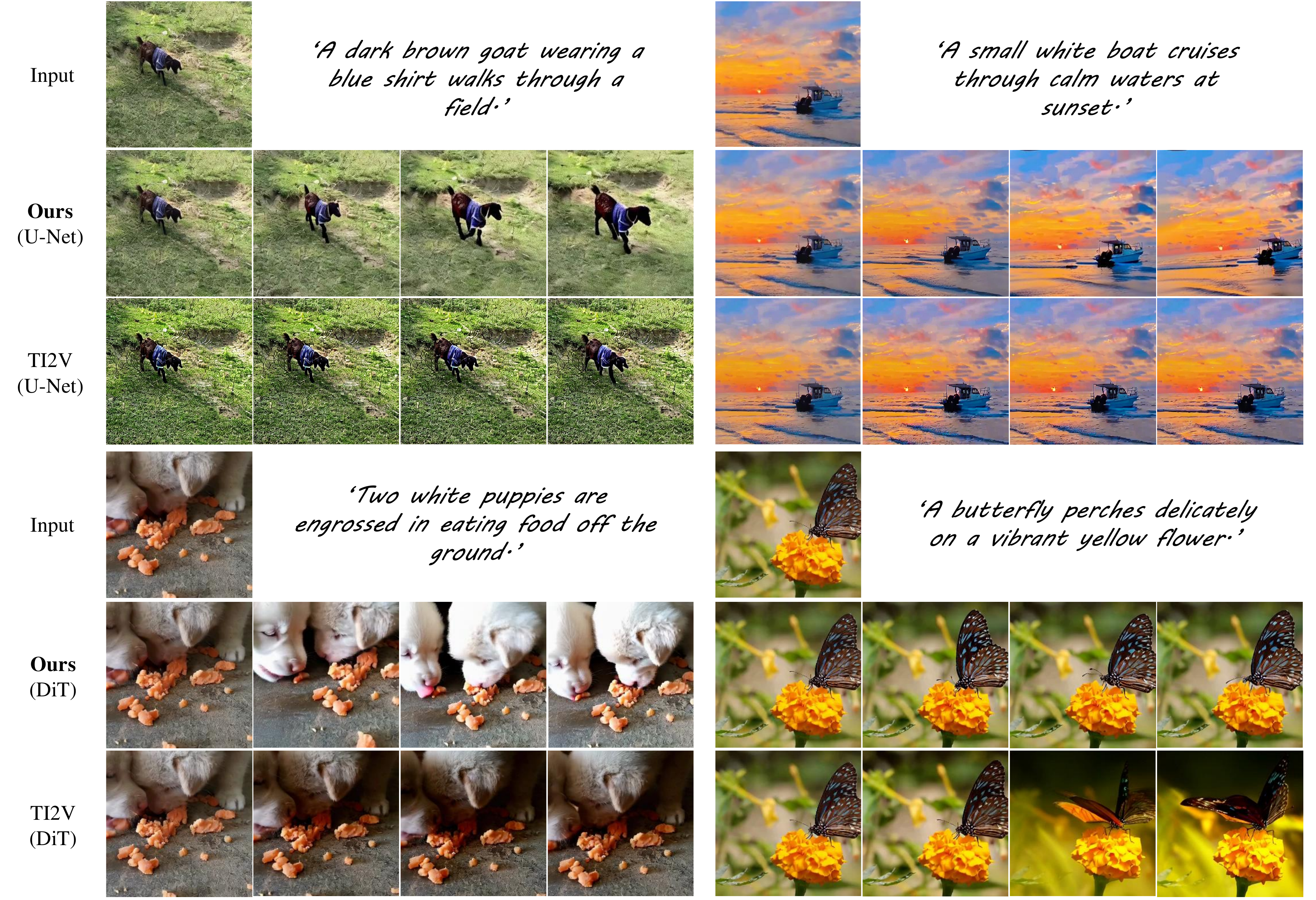}
    \caption{Qualitative comparison: Visual examples of generated videos for \method compared to the TI2V baseline on examples from the \benchmark benchmark.}
    \vspace{-0.5cm}
    \label{fig:comparison}
\end{figure*}

\subsection{Baseline Comparisons}
\label{sec:comparison}
For comparison with U-Net-based models, we evaluate our method against several open-sourced state-of-the-art image-to-video models: VideoCrafter~\citep{chen2023videocrafter1opendiffusionmodels}, DynamiCrafter~\citep{xing2023dynamicrafteranimatingopendomainimages}, Motion-I2V~\citep{shi2024motioni2vconsistentcontrollableimagetovideo}, and ConsistI2V~\citep{ren2024consisti2venhancingvisualconsistency}. 

For both U-Net and DiT models, we also report results for a single-step image-to-video baseline, denoted as TI2V, which is a variant of the proposed architecture with two main differences: (i) TI2V accepts a concatenation of the first frame and input noise as input, without any motion trajectory representation, and (ii) instead of our proposed masked attention blocks, TI2V includes additional standard attention layers to match the parameter count. Specifically, TI2V includes additional self-attention and cross-attention layers that attend to the full video patches without masking and with text prompt, respectively.

Tab.~\ref{tab:sav_benchmark} depicts our comparison on the \benchmark benchmark for both single-object and multi-object. As can be seen, the proposed method outperforms all of the evaluated baselines considering both single- and multiple object setups,
while holding comparable average displacement. Furthermore, we demonstrate significant enhancements in ViCLIP-V and FVD, underscoring its ability to generate videos that are both image-faithful and highly realistic. 
Additionally, human raters show a preference for our method over the TI2V baseline. 
Similarly, in the DiT-based models, \method exhibits clear superiority over the TI2V baseline across all evaluated metrics in both single- and multi-object settings. Notably, we achieve a significant improvement in ViCLIP-V, demonstrating substantially higher image fidelity.

Next, we report results on Image-Animation-Bench in Tab.~\ref{tab:i2v_bench}. Consistent with the findings on the \benchmark benchmark, \method surpasses all baselines on Image-Animation-Bench. The U-Net and DiT variants of \method achieve lower FVD and higher ViCLIP-T and ViCLIP-V scores. For human evaluation, we randomly sample 128 videos from the benchmark.
Human evaluations further support these findings, with \method receiving the highest preference rates across almost all comparison metrics—winning against all baselines except in text alignment with TI2V, where TI2V slightly outperforms \method (50.8 vs. 49.2). However, despite this slight outperformance by TI2V in human evaluations, \method still demonstrates superior text-alignment according to automatic metrics. 
To further illustrate the results, Fig.~\ref{fig:comparison} presents four visual examples comparing our model against TI2V, showcasing generated video examples across the \benchmark benchmark. These examples highlight superiority in motion (top left), aesthetics (top right), text alignment (bottom left), and visual consistency (bottom right). Additional samples and qualitative comparisons are available, comparisons with DiT-based models are shown in Fig.\ref{fig:comp_dit}, while comparisons with U-Net-based models are presented in Fig.~\ref{fig:comp_unet}.

\subsection{Ablation Study}
\begin{table}[t]
  \centering
  \scalebox{0.78}{
  \begin{tabular}{@{}lccccc@{}}
    \toprule
    Config. & FVD\(\downarrow\) & CF\(\uparrow\) & ViCLIP-T\(\uparrow\) & ViCLIP-V\(\uparrow\) & AD \\
    \midrule
    TI2V~(UNet) & 974.07 & 0.942 & 0.218 & 0.880 & 6.91 \\
    no mask attn~(UNet) & 972.25 & 0.962 & 0.214 & 0.880 & 4.99 \\
    w. cross-attn~(UNet) & 670.92 & 0.965 & 0.220 & 0.890 & 5.25 \\
    w. self-attn~(UNet) & 658.92 & 0.968 & 0.218 & 0.892 & 5.00 \\
    Ours~(UNet) & \textbf{648.59} & \textbf{0.968} & \textbf{0.220} & \textbf{0.892} & 5.15 \\
    \midrule
    TI2V~(DiT) & 1199.86 & 0.921 & 0.222 & 0.802 & 10.70 \\
    no mask attn~(DiT) & 1182.49 & 0.943 & 0.223 & 0.851 & 6.78 \\
    w. cross-attn~(DiT) & 1105.91 & 0.945 & 0.226 & 0.859 & 7.23 \\
    w. self-attn~(DiT) & 1152.38 & 0.946 & 0.223 & 0.855 & 7.01 \\
    Ours~(DiT) & \textbf{1082.23} & \textbf{0.947} & \textbf{0.226} & \textbf{0.861} & 7.35 \\
    \bottomrule
  \end{tabular}}
    \caption{Ablation study results on the \benchmark benchmark comparing the performance of different attention configurations, in both U-Net and DiT-based models.}
    \label{tab:ablation_masked_attention}
    \vspace{-0.5cm}
\end{table}

\myparagraph{Effect of The Masked Attention Mechanism.}
We evaluate the impact of spatial and temporal masked attention layers in \method. 
We consider the following configurations: (i) without masked attention~(no mask attn), (ii) with only masked cross-attention~(w. cross-attn), (iii) with only masked self-attention~(w. self-attn), and (iv) with both masked cross and self-attention~(\method).
We also compare against the TI2V baseline model that includes additional attention layers positioned as our masked layers, but performs self-attention and cross-attention without masking, ensuring it has the same number of parameters as \method.
Tab.~\ref{tab:ablation_masked_attention} reports the results of the ablation study on the \benchmark benchmark using both U-Net based and DiT-based models. 
Results indicate that the addition of masked attention significantly improves performance across all metrics, particularly when compared to the baseline with the same architecture but without masking. 
A qualitative comparison between the variants evaluated in this ablation study, is available in Fig.~\ref{fig:ablation_masked_attention}.

\begin{figure}[ht]
    \centering
    \includegraphics[width=0.45\textwidth]{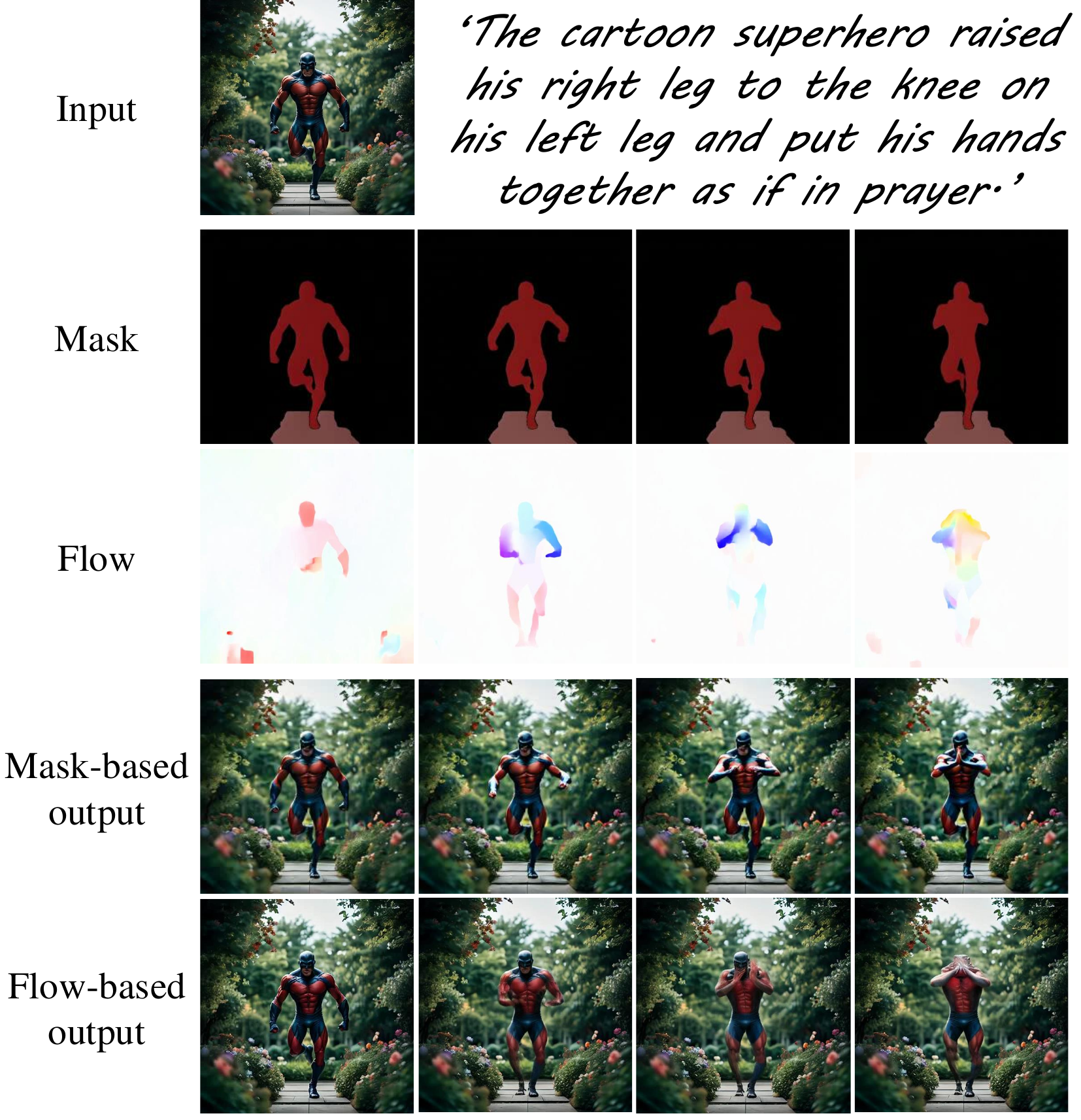}
    \caption{Qualitative comparison of generated videos using segmentation masks vs optical flow as an intermediate motion representation.
    The first row shows the input image and text, the second row displays the generated masks, and the third row presents the generated optical flow. The fourth and fifth rows show the generated videos, with the fourth row using our mask-based model and the fifth using our flow-based model.}
    \label{fig:qualitative_comparison}
    \vspace{-0.3cm}
\end{figure}

\myparagraph{Motion Representation Ablation.}
Lastly, we compare a mask-based trajectory versus optical flow based motion trajectory.
We train separate models for Stage 1 and Stage 2 of \method for each of the trajectories, keeping all other aspects the same. 
Tab.~\ref{tab:seg_vs_flow} reports the results using the U-Net variant of our method. 
As can be seen, using segmentation masks significantly outperforms the optical flow.
We hypothesize that optical flow is too restrictive for this task, as it enforces precise pixel-wise correspondences and tends to collapse, as shown in the bottom right frame of Fig.~\ref{fig:qualitative_comparison}, which provides a qualitative illustration of this ablation. Another qualitative example is shown in Fig.\ref{fig:vs_flow}.
This rigidity leads Stage 2 to overly rely on the provided motion, limiting its ability to generate realistic and detailed video content.
In contrast, segmentation masks offer a higher-level semantic representation of object motion, providing guidance without constraining the model to exact pixel movements as demonstrated by SpaText~\cite{avrahami2023spatext}~(see Sec~.4.3.).
The latter allows Stage 2 to follow the motion cues while autonomously refining fine details, resulting in more natural and coherent video sequences.

\begin{table}[t]
    \centering
    \scalebox{0.9}{
    \begin{tabular}{@{}lccccc@{}}
        \toprule
        Configuration & FVD$\downarrow$ & CF$\uparrow$ & ViCLIP-T$\uparrow$ & ViCLIP-V$\uparrow$ & AD \\
        \midrule
        w. OF & 1014.72 & 0.934 & 0.219 & 0.879 & 7.04 \\
        w. Seg. & \textbf{648.59} & \textbf{0.968} & \textbf{0.220} & \textbf{0.892} & 5.15 \\
        \bottomrule
    \end{tabular}}
    \caption{Ablation study comparing segmentation masks and optical flow as motion trajectory representation. \textbf{w. Seg} refers to models with segmentation-based motion trajectories, while \textbf{w. OF} denotes models with optical flow-based motion trajectories.}
    \label{tab:seg_vs_flow}
    \vspace{-0.3cm}
\end{table}

\section{Conclusion}

We presented \method, a novel two-stage framework for image-to-video generation leveraging mask-based motion trajectories as an intermediate representation, enabling coherent and realistic multi-object motion in generated videos. Motion trajectories are injected via two attention-based objectives that effectively use this representation to enforce the predicted motion and semantics in the generated video. Our approach empirically achieves SOTA performance in challenging single- and multi-object settings.

{
    \small
    \bibliographystyle{ieeenat_fullname}
    \bibliography{ref}

\begin{thebibliography}{56}
\providecommand{\natexlab}[1]{#1}
\providecommand{\url}[1]{\texttt{#1}}
\expandafter\ifx\csname urlstyle\endcsname\relax
  \providecommand{\doi}[1]{doi: #1}\else
  \providecommand{\doi}{doi: \begingroup \urlstyle{rm}\Url}\fi

\bibitem[Albergo et~al.(2023)Albergo, Boffi, and Vanden-Eijnden]{albergo2023stochastic}
Michael~S Albergo, Nicholas~M Boffi, and Eric Vanden-Eijnden.
\newblock Stochastic interpolants: A unifying framework for flows and diffusions.
\newblock \emph{arXiv preprint arXiv:2303.08797}, 2023.

\bibitem[An et~al.(2023)An, Zhang, Yang, Gupta, Huang, Luo, and Yin]{an2023latent}
Jie An, Songyang Zhang, Harry Yang, Sonal Gupta, Jia-Bin Huang, Jiebo Luo, and Xi Yin.
\newblock Latent-shift: Latent diffusion with temporal shift for efficient text-to-video generation.
\newblock \emph{arXiv preprint arXiv:2304.08477}, 2023.

\bibitem[Avrahami et~al.(2023)Avrahami, Hayes, Gafni, Gupta, Taigman, Parikh, Lischinski, Fried, and Yin]{avrahami2023spatext}
Omri Avrahami, Thomas Hayes, Oran Gafni, Sonal Gupta, Yaniv Taigman, Devi Parikh, Dani Lischinski, Ohad Fried, and Xi Yin.
\newblock Spatext: Spatio-textual representation for controllable image generation.
\newblock In \emph{Proceedings of the IEEE/CVF Conference on Computer Vision and Pattern Recognition}, pages 18370--18380, 2023.

\bibitem[Betker et~al.(2023)Betker, Goh, Jing, Brooks, Wang, Li, Ouyang, Zhuang, Lee, Guo, et~al.]{dalle3}
James Betker, Gabriel Goh, Li Jing, Tim Brooks, Jianfeng Wang, Linjie Li, Long Ouyang, Juntang Zhuang, Joyce Lee, Yufei Guo, et~al.
\newblock Improving image generation with better captions.
\newblock \emph{Computer Science. https://cdn. openai. com/papers/dall-e-3. pdf}, 2\penalty0 (3):\penalty0 8, 2023.

\bibitem[Blattmann et~al.(2021)Blattmann, Milbich, Dorkenwald, and Ommer]{blattmann2021understanding}
Andreas Blattmann, Timo Milbich, Michael Dorkenwald, and Bjorn Ommer.
\newblock Understanding object dynamics for interactive image-to-video synthesis.
\newblock In \emph{Proceedings of the IEEE/CVF Conference on Computer Vision and Pattern Recognition}, pages 5171--5181, 2021.

\bibitem[Blattmann et~al.(2023{\natexlab{a}})Blattmann, Dockhorn, Kulal, Mendelevitch, Kilian, Lorenz, Levi, English, Voleti, Letts, Jampani, and Rombach]{blattmann2023stablevideodiffusionscaling}
Andreas Blattmann, Tim Dockhorn, Sumith Kulal, Daniel Mendelevitch, Maciej Kilian, Dominik Lorenz, Yam Levi, Zion English, Vikram Voleti, Adam Letts, Varun Jampani, and Robin Rombach.
\newblock Stable video diffusion: Scaling latent video diffusion models to large datasets, 2023{\natexlab{a}}.

\bibitem[Blattmann et~al.(2023{\natexlab{b}})Blattmann, Rombach, Ling, Dockhorn, Kim, Fidler, and Kreis]{blattmann2023align}
Andreas Blattmann, Robin Rombach, Huan Ling, Tim Dockhorn, Seung~Wook Kim, Sanja Fidler, and Karsten Kreis.
\newblock Align your latents: High-resolution video synthesis with latent diffusion models.
\newblock In \emph{Proceedings of the IEEE/CVF Conference on Computer Vision and Pattern Recognition}, pages 22563--22575, 2023{\natexlab{b}}.

\bibitem[Brooks et~al.(2023)]{brooks2023instructpix2pix}
Tim Brooks et~al.
\newblock Instructpix2pix: Learning to follow image editing instructions.
\newblock In \emph{CVPR}, 2023.

\bibitem[Chen et~al.(2023)Chen, Xia, He, Zhang, Cun, Yang, Xing, Liu, Chen, Wang, Weng, and Shan]{chen2023videocrafter1opendiffusionmodels}
Haoxin Chen, Menghan Xia, Yingqing He, Yong Zhang, Xiaodong Cun, Shaoshu Yang, Jinbo Xing, Yaofang Liu, Qifeng Chen, Xintao Wang, Chao Weng, and Ying Shan.
\newblock Videocrafter1: Open diffusion models for high-quality video generation, 2023.

\bibitem[Chen et~al.(2024)Chen, Zhang, Cun, Xia, Wang, Weng, and Shan]{chen2024videocrafter2}
Haoxin Chen, Yong Zhang, Xiaodong Cun, Menghan Xia, Xintao Wang, Chao Weng, and Ying Shan.
\newblock Videocrafter2: Overcoming data limitations for high-quality video diffusion models.
\newblock In \emph{Proceedings of the IEEE/CVF Conference on Computer Vision and Pattern Recognition}, pages 7310--7320, 2024.

\bibitem[Dai et~al.(2023{\natexlab{a}})Dai, Hou, Ma, Tsai, Wang, Wang, Zhang, Vandenhende, Wang, Dubey, et~al.]{emu}
Xiaoliang Dai, Ji Hou, Chih-Yao Ma, Sam Tsai, Jialiang Wang, Rui Wang, Peizhao Zhang, Simon Vandenhende, Xiaofang Wang, Abhimanyu Dubey, et~al.
\newblock Emu: Enhancing image generation models using photogenic needles in a haystack.
\newblock \emph{arXiv preprint arXiv:2309.15807}, 2023{\natexlab{a}}.

\bibitem[Dai et~al.(2023{\natexlab{b}})Dai, Zhang, Yao, Qiu, Zhu, Qin, and Wang]{dai2023animateanythingfinegrainedopendomain}
Zuozhuo Dai, Zhenghao Zhang, Yao Yao, Bingxue Qiu, Siyu Zhu, Long Qin, and Weizhi Wang.
\newblock Animateanything: Fine-grained open domain image animation with motion guidance, 2023{\natexlab{b}}.

\bibitem[Dubey et~al.(2024)]{dubey2024llama3herdmodels}
Abhimanyu Dubey et~al.
\newblock The llama 3 herd of models, 2024.

\bibitem[Esser et~al.(2024)Esser, Kulal, Blattmann, Entezari, M{\"u}ller, Saini, Levi, Lorenz, Sauer, Boesel, et~al.]{stable3}
Patrick Esser, Sumith Kulal, Andreas Blattmann, Rahim Entezari, Jonas M{\"u}ller, Harry Saini, Yam Levi, Dominik Lorenz, Axel Sauer, Frederic Boesel, et~al.
\newblock Scaling rectified flow transformers for high-resolution image synthesis.
\newblock In \emph{Forty-first International Conference on Machine Learning}, 2024.

\bibitem[Ge et~al.(2023)Ge, Nah, Liu, Poon, Tao, Catanzaro, Jacobs, Huang, Liu, and Balaji]{ge2023preserve}
Songwei Ge, Seungjun Nah, Guilin Liu, Tyler Poon, Andrew Tao, Bryan Catanzaro, David Jacobs, Jia-Bin Huang, Ming-Yu Liu, and Yogesh Balaji.
\newblock Preserve your own correlation: A noise prior for video diffusion models.
\newblock In \emph{Proceedings of the IEEE/CVF International Conference on Computer Vision}, pages 22930--22941, 2023.

\bibitem[Girdhar et~al.(2023)Girdhar, Singh, Brown, Duval, Azadi, Rambhatla, Shah, Yin, Parikh, and Misra]{girdhar2023emu}
Rohit Girdhar, Mannat Singh, Andrew Brown, Quentin Duval, Samaneh Azadi, Sai~Saketh Rambhatla, Akbar Shah, Xi Yin, Devi Parikh, and Ishan Misra.
\newblock Emu video: Factorizing text-to-video generation by explicit image conditioning.
\newblock \emph{arXiv preprint arXiv:2311.10709}, 2023.

\bibitem[Guo et~al.(2024)Guo, Yang, Rao, Liang, Wang, Qiao, Agrawala, Lin, and Dai]{guo2024animatediffanimatepersonalizedtexttoimage}
Yuwei Guo, Ceyuan Yang, Anyi Rao, Zhengyang Liang, Yaohui Wang, Yu Qiao, Maneesh Agrawala, Dahua Lin, and Bo Dai.
\newblock Animatediff: Animate your personalized text-to-image diffusion models without specific tuning, 2024.

\bibitem[He et~al.(2022)He, Yang, Zhang, Shan, and Chen]{he2022latent}
Yingqing He, Tianyu Yang, Yong Zhang, Ying Shan, and Qifeng Chen.
\newblock Latent video diffusion models for high-fidelity long video generation.
\newblock \emph{arXiv preprint arXiv:2211.13221}, 2022.

\bibitem[Ho and Salimans(2022)]{ho2022classifier}
Jonathan Ho and Tim Salimans.
\newblock Classifier-free diffusion guidance.
\newblock \emph{arXiv preprint arXiv:2207.12598}, 2022.

\bibitem[Ho et~al.(2020)Ho, Jain, and Abbeel]{ho2020denoising}
Jonathan Ho, Ajay Jain, and Pieter Abbeel.
\newblock Denoising diffusion probabilistic models.
\newblock \emph{Advances in neural information processing systems}, 33:\penalty0 6840--6851, 2020.

\bibitem[Ho et~al.(2022{\natexlab{a}})Ho, Chan, Saharia, Whang, Gao, Gritsenko, Kingma, Poole, Norouzi, Fleet, et~al.]{ho2022imagen}
Jonathan Ho, William Chan, Chitwan Saharia, Jay Whang, Ruiqi Gao, Alexey Gritsenko, Diederik~P Kingma, Ben Poole, Mohammad Norouzi, David~J Fleet, et~al.
\newblock Imagen video: High definition video generation with diffusion models.
\newblock \emph{arXiv preprint arXiv:2210.02303}, 2022{\natexlab{a}}.

\bibitem[Ho et~al.(2022{\natexlab{b}})Ho, Salimans, Gritsenko, Chan, Norouzi, and Fleet]{ho2022video}
Jonathan Ho, Tim Salimans, Alexey Gritsenko, William Chan, Mohammad Norouzi, and David~J Fleet.
\newblock Video diffusion models.
\newblock \emph{Advances in Neural Information Processing Systems}, 35:\penalty0 8633--8646, 2022{\natexlab{b}}.

\bibitem[Kingma(2013)]{kingma2013auto}
Diederik~P Kingma.
\newblock Auto-encoding variational bayes.
\newblock \emph{arXiv preprint arXiv:1312.6114}, 2013.

\bibitem[Li et~al.(2023{\natexlab{a}})Li, Chu, Wu, Yuan, Liu, Zhang, Li, Feng, Ding, and Wang]{li2023videogen}
Xin Li, Wenqing Chu, Ye Wu, Weihang Yuan, Fanglong Liu, Qi Zhang, Fu Li, Haocheng Feng, Errui Ding, and Jingdong Wang.
\newblock Videogen: A reference-guided latent diffusion approach for high definition text-to-video generation.
\newblock \emph{arXiv preprint arXiv:2309.00398}, 2023{\natexlab{a}}.

\bibitem[Li et~al.(2023{\natexlab{b}})Li, Liu, Wu, Mu, Yang, Gao, Li, and Lee]{li2023gligenopensetgroundedtexttoimage}
Yuheng Li, Haotian Liu, Qingyang Wu, Fangzhou Mu, Jianwei Yang, Jianfeng Gao, Chunyuan Li, and Yong~Jae Lee.
\newblock Gligen: Open-set grounded text-to-image generation, 2023{\natexlab{b}}.

\bibitem[Lin et~al.(2024)Lin, Liu, Li, and Yang]{lin2024commondiffusionnoiseschedules}
Shanchuan Lin, Bingchen Liu, Jiashi Li, and Xiao Yang.
\newblock Common diffusion noise schedules and sample steps are flawed, 2024.

\bibitem[Lipman et~al.(2023)Lipman, Chen, Ben-Hamu, Nickel, and Le]{lipman2023flowmatchinggenerativemodeling}
Yaron Lipman, Ricky T.~Q. Chen, Heli Ben-Hamu, Maximilian Nickel, and Matt Le.
\newblock Flow matching for generative modeling, 2023.

\bibitem[Liu(2022)]{liu2022rectified}
Qiang Liu.
\newblock Rectified flow: A marginal preserving approach to optimal transport.
\newblock \emph{arXiv preprint arXiv:2209.14577}, 2022.

\bibitem[Liu et~al.(2024)Liu, Zeng, Ren, Li, Zhang, Yang, Jiang, Li, Yang, Su, Zhu, and Zhang]{liu2024groundingdinomarryingdino}
Shilong Liu, Zhaoyang Zeng, Tianhe Ren, Feng Li, Hao Zhang, Jie Yang, Qing Jiang, Chunyuan Li, Jianwei Yang, Hang Su, Jun Zhu, and Lei Zhang.
\newblock Grounding dino: Marrying dino with grounded pre-training for open-set object detection, 2024.

\bibitem[Ma et~al.(2024{\natexlab{a}})Ma, Wang, Jia, Chen, Li, Chen, and Qiao]{ma2024cinemoconsistentcontrollableimage}
Xin Ma, Yaohui Wang, Gengyun Jia, Xinyuan Chen, Yuan-Fang Li, Cunjian Chen, and Yu Qiao.
\newblock Cinemo: Consistent and controllable image animation with motion diffusion models, 2024{\natexlab{a}}.

\bibitem[Ma et~al.(2024{\natexlab{b}})Ma, Wang, Jia, Chen, Liu, Li, Chen, and Qiao]{ma2024latte}
Xin Ma, Yaohui Wang, Gengyun Jia, Xinyuan Chen, Ziwei Liu, Yuan-Fang Li, Cunjian Chen, and Yu Qiao.
\newblock Latte: Latent diffusion transformer for video generation.
\newblock \emph{arXiv preprint arXiv:2401.03048}, 2024{\natexlab{b}}.

\bibitem[Menapace et~al.(2024)Menapace, Siarohin, Skorokhodov, Deyneka, Chen, Kag, Fang, Stoliar, Ricci, Ren, et~al.]{menapace2024snap}
Willi Menapace, Aliaksandr Siarohin, Ivan Skorokhodov, Ekaterina Deyneka, Tsai-Shien Chen, Anil Kag, Yuwei Fang, Aleksei Stoliar, Elisa Ricci, Jian Ren, et~al.
\newblock Snap video: Scaled spatiotemporal transformers for text-to-video synthesis.
\newblock In \emph{Proceedings of the IEEE/CVF Conference on Computer Vision and Pattern Recognition}, pages 7038--7048, 2024.

\bibitem[Nie et~al.(2024)Nie, Liu, Mardani, Liu, Eckart, and Vahdat]{nie2024compositionaltexttoimagegenerationdense}
Weili Nie, Sifei Liu, Morteza Mardani, Chao Liu, Benjamin Eckart, and Arash Vahdat.
\newblock Compositional text-to-image generation with dense blob representations, 2024.

\bibitem[OpenAI(2024)]{sora}
OpenAI.
\newblock Video generation models as world simulators.
\newblock \url{https://openai.com/index/video-generation-models-as-world-simulators/}, 2024.

\bibitem[Pan et~al.(2019)Pan, Wang, Jia, Shao, Sheng, Yan, and Wang]{pan2019video}
Junting Pan, Chengyu Wang, Xu Jia, Jing Shao, Lu Sheng, Junjie Yan, and Xiaogang Wang.
\newblock Video generation from single semantic label map.
\newblock In \emph{Proceedings of the IEEE/CVF Conference on Computer Vision and Pattern Recognition}, pages 3733--3742, 2019.

\bibitem[Peebles and Xie(2023)]{peebles2023scalable}
William Peebles and Saining Xie.
\newblock Scalable diffusion models with transformers.
\newblock In \emph{Proceedings of the IEEE/CVF International Conference on Computer Vision}, pages 4195--4205, 2023.

\bibitem[Podell et~al.(2023)Podell, English, Lacey, Blattmann, Dockhorn, M{\"u}ller, Penna, and Rombach]{podell2023sdxl}
Dustin Podell, Zion English, Kyle Lacey, Andreas Blattmann, Tim Dockhorn, Jonas M{\"u}ller, Joe Penna, and Robin Rombach.
\newblock Sdxl: Improving latent diffusion models for high-resolution image synthesis.
\newblock \emph{arXiv preprint arXiv:2307.01952}, 2023.

\bibitem[Polyak et~al.(2024)Polyak, Zohar, Brown, Tjandra, Sinha, Lee, Vyas, Shi, Ma, Chuang, Yan, Choudhary, Wang, Sethi, Pang, Ma, Misra, Hou, Wang, Jagadeesh, Li, Zhang, Singh, Williamson, Le, Yu, Singh, Zhang, Vajda, Duval, Girdhar, Sumbaly, Rambhatla, Tsai, Azadi, Datta, Chen, Bell, Ramaswamy, Sheynin, Bhattacharya, Motwani, Xu, Li, Hou, Hsu, Yin, Dai, Taigman, Luo, Liu, Wu, Zhao, Kirstain, He, He, Pumarola, Thabet, Sanakoyeu, Mallya, Guo, Araya, Kerr, Wood, Liu, Peng, Vengertsev, Schonfeld, Blanchard, Juefei-Xu, Nord, Liang, Hoffman, Kohler, Fire, Sivakumar, Chen, Yu, Gao, Georgopoulos, Moritz, Sampson, Li, Parmeggiani, Fine, Fowler, Petrovic, and Du]{polyak2024moviegencastmedia}
Adam Polyak, Amit Zohar, Andrew Brown, Andros Tjandra, Animesh Sinha, Ann Lee, Apoorv Vyas, Bowen Shi, Chih-Yao Ma, Ching-Yao Chuang, David Yan, Dhruv Choudhary, Dingkang Wang, Geet Sethi, Guan Pang, Haoyu Ma, Ishan Misra, Ji Hou, Jialiang Wang, Kiran Jagadeesh, Kunpeng Li, Luxin Zhang, Mannat Singh, Mary Williamson, Matt Le, Matthew Yu, Mitesh~Kumar Singh, Peizhao Zhang, Peter Vajda, Quentin Duval, Rohit Girdhar, Roshan Sumbaly, Sai~Saketh Rambhatla, Sam Tsai, Samaneh Azadi, Samyak Datta, Sanyuan Chen, Sean Bell, Sharadh Ramaswamy, Shelly Sheynin, Siddharth Bhattacharya, Simran Motwani, Tao Xu, Tianhe Li, Tingbo Hou, Wei-Ning Hsu, Xi Yin, Xiaoliang Dai, Yaniv Taigman, Yaqiao Luo, Yen-Cheng Liu, Yi-Chiao Wu, Yue Zhao, Yuval Kirstain, Zecheng He, Zijian He, Albert Pumarola, Ali Thabet, Artsiom Sanakoyeu, Arun Mallya, Baishan Guo, Boris Araya, Breena Kerr, Carleigh Wood, Ce Liu, Cen Peng, Dimitry Vengertsev, Edgar Schonfeld, Elliot Blanchard, Felix Juefei-Xu, Fraylie Nord, Jeff Liang, John Hoffman, Jonas
  Kohler, Kaolin Fire, Karthik Sivakumar, Lawrence Chen, Licheng Yu, Luya Gao, Markos Georgopoulos, Rashel Moritz, Sara~K. Sampson, Shikai Li, Simone Parmeggiani, Steve Fine, Tara Fowler, Vladan Petrovic, and Yuming Du.
\newblock Movie gen: A cast of media foundation models, 2024.

\bibitem[Radford et~al.(2021)Radford, Kim, Hallacy, Ramesh, Goh, Agarwal, Sastry, Askell, Mishkin, Clark, Krueger, and Sutskever]{radford2021learningtransferablevisualmodels}
Alec Radford, Jong~Wook Kim, Chris Hallacy, Aditya Ramesh, Gabriel Goh, Sandhini Agarwal, Girish Sastry, Amanda Askell, Pamela Mishkin, Jack Clark, Gretchen Krueger, and Ilya Sutskever.
\newblock Learning transferable visual models from natural language supervision, 2021.

\bibitem[Ravi et~al.(2024)Ravi, Gabeur, Hu, Hu, Ryali, Ma, Khedr, Rädle, Rolland, Gustafson, Mintun, Pan, Alwala, Carion, Wu, Girshick, Dollár, and Feichtenhofer]{ravi2024sam2segmentimages}
Nikhila Ravi, Valentin Gabeur, Yuan-Ting Hu, Ronghang Hu, Chaitanya Ryali, Tengyu Ma, Haitham Khedr, Roman Rädle, Chloe Rolland, Laura Gustafson, Eric Mintun, Junting Pan, Kalyan~Vasudev Alwala, Nicolas Carion, Chao-Yuan Wu, Ross Girshick, Piotr Dollár, and Christoph Feichtenhofer.
\newblock Sam 2: Segment anything in images and videos, 2024.

\bibitem[Ren et~al.(2024)Ren, Yang, Zhang, Wei, Du, Huang, and Chen]{ren2024consisti2venhancingvisualconsistency}
Weiming Ren, Huan Yang, Ge Zhang, Cong Wei, Xinrun Du, Wenhao Huang, and Wenhu Chen.
\newblock Consisti2v: Enhancing visual consistency for image-to-video generation, 2024.

\bibitem[Rombach et~al.(2022)Rombach, Blattmann, Lorenz, Esser, and Ommer]{rombach2022high}
Robin Rombach, Andreas Blattmann, Dominik Lorenz, Patrick Esser, and Bj{\"o}rn Ommer.
\newblock High-resolution image synthesis with latent diffusion models.
\newblock In \emph{Proceedings of the IEEE/CVF conference on computer vision and pattern recognition}, pages 10684--10695, 2022.

\bibitem[Shi et~al.(2024)Shi, Huang, Wang, Bian, Li, Zhang, Zhang, Cheung, See, Qin, Dai, and Li]{shi2024motioni2vconsistentcontrollableimagetovideo}
Xiaoyu Shi, Zhaoyang Huang, Fu-Yun Wang, Weikang Bian, Dasong Li, Yi Zhang, Manyuan Zhang, Ka~Chun Cheung, Simon See, Hongwei Qin, Jifeng Dai, and Hongsheng Li.
\newblock Motion-i2v: Consistent and controllable image-to-video generation with explicit motion modeling, 2024.

\bibitem[Singer et~al.(2022)Singer, Polyak, Hayes, Yin, An, Zhang, Hu, Yang, Ashual, Gafni, Parikh, Gupta, and Taigman]{singer2022makeavideotexttovideogenerationtextvideo}
Uriel Singer, Adam Polyak, Thomas Hayes, Xi Yin, Jie An, Songyang Zhang, Qiyuan Hu, Harry Yang, Oron Ashual, Oran Gafni, Devi Parikh, Sonal Gupta, and Yaniv Taigman.
\newblock Make-a-video: Text-to-video generation without text-video data, 2022.

\bibitem[Song et~al.(2020{\natexlab{a}})Song, Meng, and Ermon]{song2020denoising}
Jiaming Song, Chenlin Meng, and Stefano Ermon.
\newblock Denoising diffusion implicit models.
\newblock \emph{arXiv preprint arXiv:2010.02502}, 2020{\natexlab{a}}.

\bibitem[Song et~al.(2020{\natexlab{b}})Song, Sohl-Dickstein, Kingma, Kumar, Ermon, and Poole]{song2020score}
Yang Song, Jascha Sohl-Dickstein, Diederik~P Kingma, Abhishek Kumar, Stefano Ermon, and Ben Poole.
\newblock Score-based generative modeling through stochastic differential equations.
\newblock \emph{arXiv preprint arXiv:2011.13456}, 2020{\natexlab{b}}.

\bibitem[Teed and Deng(2020)]{teed2020raftrecurrentallpairsfield}
Zachary Teed and Jia Deng.
\newblock Raft: Recurrent all-pairs field transforms for optical flow, 2020.

\bibitem[Unterthiner et~al.(2019)Unterthiner, van Steenkiste, Kurach, Marinier, Michalski, and Gelly]{unterthiner2019accurategenerativemodelsvideo}
Thomas Unterthiner, Sjoerd van Steenkiste, Karol Kurach, Raphael Marinier, Marcin Michalski, and Sylvain Gelly.
\newblock Towards accurate generative models of video: A new metric \& challenges, 2019.

\bibitem[Wang et~al.(2023{\natexlab{a}})Wang, Yuan, Chen, Zhang, Wang, and Zhang]{wang2023modelscope}
Jiuniu Wang, Hangjie Yuan, Dayou Chen, Yingya Zhang, Xiang Wang, and Shiwei Zhang.
\newblock Modelscope text-to-video technical report.
\newblock \emph{arXiv preprint arXiv:2308.06571}, 2023{\natexlab{a}}.

\bibitem[Wang et~al.(2023{\natexlab{b}})Wang, Yang, Tuo, He, Zhu, Fu, and Liu]{wang2023videofactory}
Wenjing Wang, Huan Yang, Zixi Tuo, Huiguo He, Junchen Zhu, Jianlong Fu, and Jiaying Liu.
\newblock Videofactory: Swap attention in spatiotemporal diffusions for text-to-video generation.
\newblock \emph{arXiv preprint arXiv:2305.10874}, 2023{\natexlab{b}}.

\bibitem[Wang et~al.(2023{\natexlab{c}})Wang, Chen, Ma, Zhou, Huang, Wang, Yang, He, Yu, Yang, et~al.]{wang2023lavie}
Yaohui Wang, Xinyuan Chen, Xin Ma, Shangchen Zhou, Ziqi Huang, Yi Wang, Ceyuan Yang, Yinan He, Jiashuo Yu, Peiqing Yang, et~al.
\newblock Lavie: High-quality video generation with cascaded latent diffusion models.
\newblock \emph{arXiv preprint arXiv:2309.15103}, 2023{\natexlab{c}}.

\bibitem[Wang et~al.(2024)Wang, He, Li, Li, Yu, Ma, Li, Chen, Chen, Wang, He, Luo, Liu, Wang, Wang, and Qiao]{wang2024internvidlargescalevideotextdataset}
Yi Wang, Yinan He, Yizhuo Li, Kunchang Li, Jiashuo Yu, Xin Ma, Xinhao Li, Guo Chen, Xinyuan Chen, Yaohui Wang, Conghui He, Ping Luo, Ziwei Liu, Yali Wang, Limin Wang, and Yu Qiao.
\newblock Internvid: A large-scale video-text dataset for multimodal understanding and generation, 2024.

\bibitem[Wu et~al.(2023)Wu, Li, Gao, Dong, Bai, Singh, Xiang, Li, Huang, Sun, He, Hu, Hu, Huang, Zhu, Cheng, Tang, Shou, Keutzer, and Iandola]{wu2023cvpr2023textguided}
Jay~Zhangjie Wu, Xiuyu Li, Difei Gao, Zhen Dong, Jinbin Bai, Aishani Singh, Xiaoyu Xiang, Youzeng Li, Zuwei Huang, Yuanxi Sun, Rui He, Feng Hu, Junhua Hu, Hai Huang, Hanyu Zhu, Xu Cheng, Jie Tang, Mike~Zheng Shou, Kurt Keutzer, and Forrest Iandola.
\newblock Cvpr 2023 text guided video editing competition, 2023.

\bibitem[Xing et~al.(2023)Xing, Xia, Zhang, Chen, Yu, Liu, Wang, Wong, and Shan]{xing2023dynamicrafteranimatingopendomainimages}
Jinbo Xing, Menghan Xia, Yong Zhang, Haoxin Chen, Wangbo Yu, Hanyuan Liu, Xintao Wang, Tien-Tsin Wong, and Ying Shan.
\newblock Dynamicrafter: Animating open-domain images with video diffusion priors, 2023.

\bibitem[Zhang et~al.(2023)Zhang, Wang, Zhang, Zhao, Yuan, Qin, Wang, Zhao, and Zhou]{zhang2023i2vgenxlhighqualityimagetovideosynthesis}
Shiwei Zhang, Jiayu Wang, Yingya Zhang, Kang Zhao, Hangjie Yuan, Zhiwu Qin, Xiang Wang, Deli Zhao, and Jingren Zhou.
\newblock I2vgen-xl: High-quality image-to-video synthesis via cascaded diffusion models, 2023.

\bibitem[Zhou et~al.(2022)Zhou, Wang, Yan, Lv, Zhu, and Feng]{zhou2022magicvideo}
Daquan Zhou, Weimin Wang, Hanshu Yan, Weiwei Lv, Yizhe Zhu, and Jiashi Feng.
\newblock Magicvideo: Efficient video generation with latent diffusion models.
\newblock \emph{arXiv preprint arXiv:2211.11018}, 2022.

\end{thebibliography}
}

\clearpage
\setcounter{page}{1}
\maketitlesupplementary

\section{Additional Results}

\subsection{Qualitative Comparison of Masked Attention Mechanism}
\label{sec:ablation_masked_attention}
\begin{figure*}[]
  \centering
    \includegraphics[width=1\textwidth]{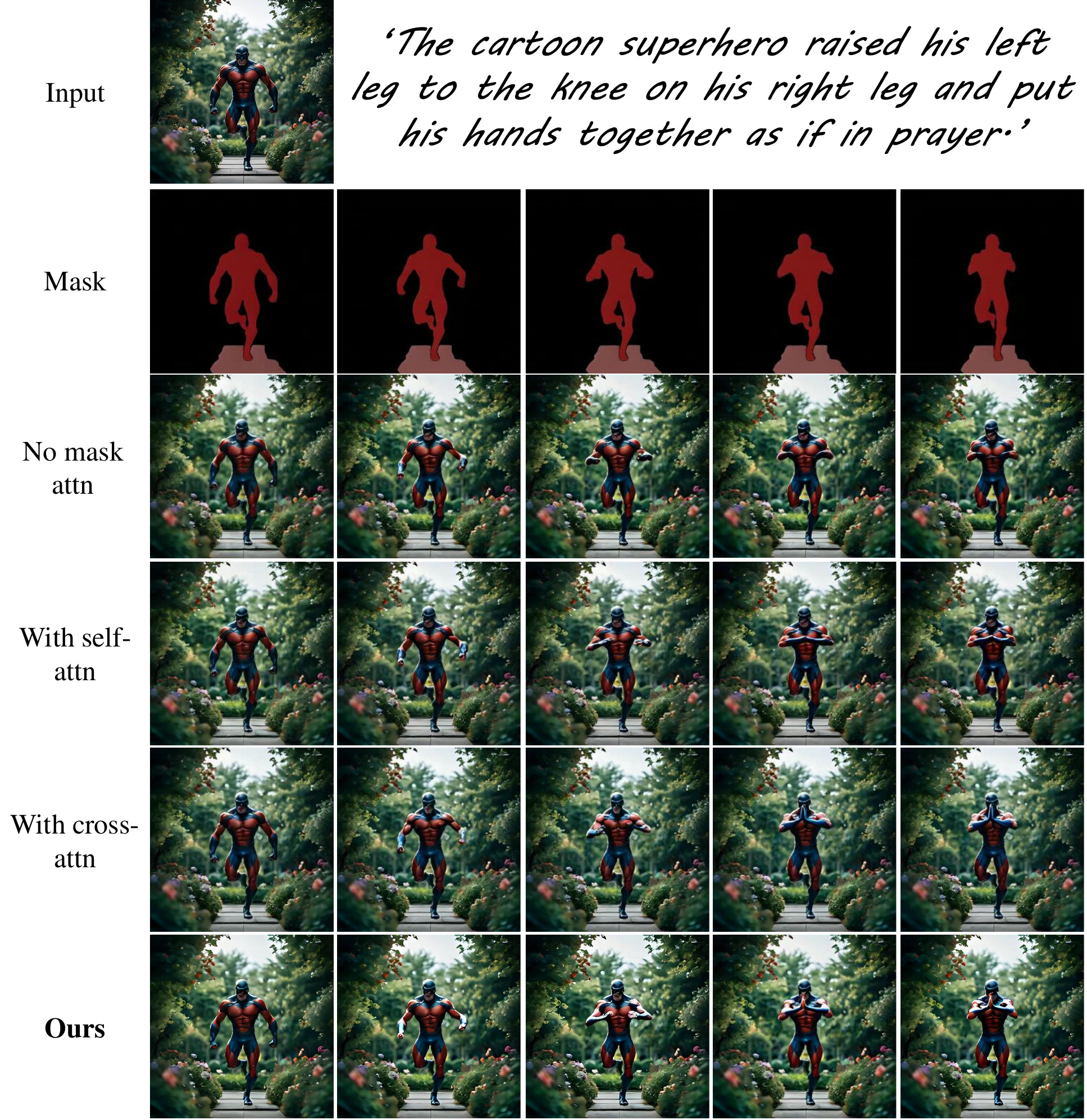}
    \caption{Qualitative comparison of generated videos for each configuration of \method. The results highlight differences when applying masked cross-attention (With cross-attn), self-attention (With self-attn), both (Ours), or no masked attention layers (No mask attn). Without masked attention, the cartoon superhero fails to perform a prayer. With masked self-attention, the superhero also fails, but the movement appears smoother and more consistent. With masked cross-attention, the superhero successfully performs the prayer, though his fingers turn blue. When integrating the full masked attention mechanism, the superhero performs the action correctly.}
    \label{fig:ablation_masked_attention}
\end{figure*}

\cref{fig:ablation_masked_attention} shows qualitative comparison of generated videos for each configuration of \method, demonstrating the differences when applying masked cross-attention, self-attention, both, or no masked attention layers.

\subsection{Qualitative Comparison of Motion Representation Ablation} \label{sec:ablation_representation} 
\begin{figure*}[]
  \centering
    \includegraphics[width=\textwidth]{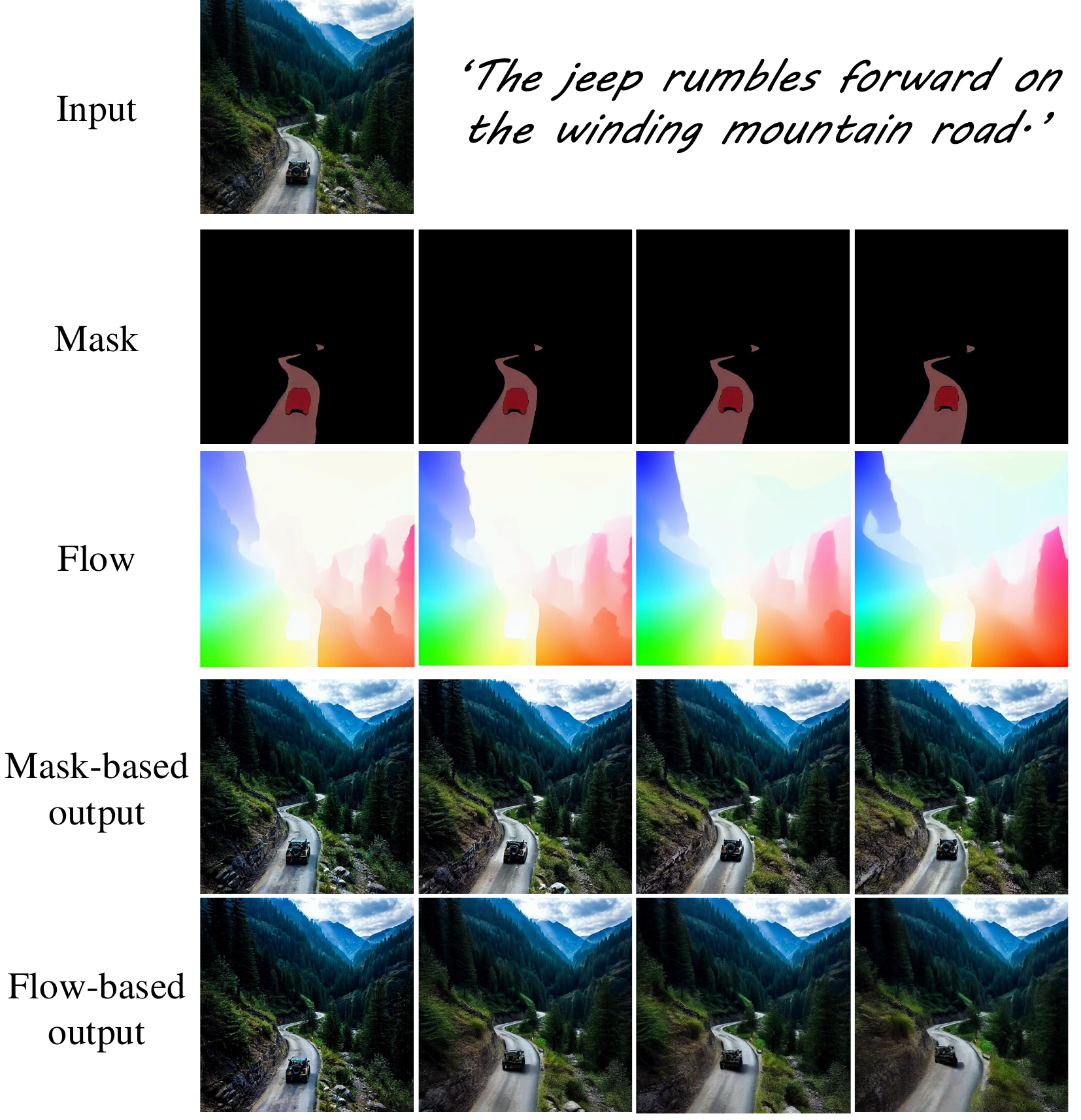}
    \caption{Qualitative comparison of generated videos using segmentation masks vs optical flow as an intermediate motion representation. The first row shows the input image and text, the second row displays the generated masks, and the third row presents the generated optical flow. The fourth and fifth rows show the generated videos, with the fourth row using our mask-based model and
    the fifth using our flow-based model.}
    \label{fig:vs_flow}
\end{figure*}

\cref{fig:vs_flow} shows a qualitative comparison of the generated videos for different intermediate representation configurations of \method. Specifically, it compares our chosen representation, which is mask-based motion trajectories, to optical flow.

\subsection{Additional Qualitative Comparisons of DiT Architecture} 
\begin{figure*}[]
  \centering
    \includegraphics[width=\textwidth]{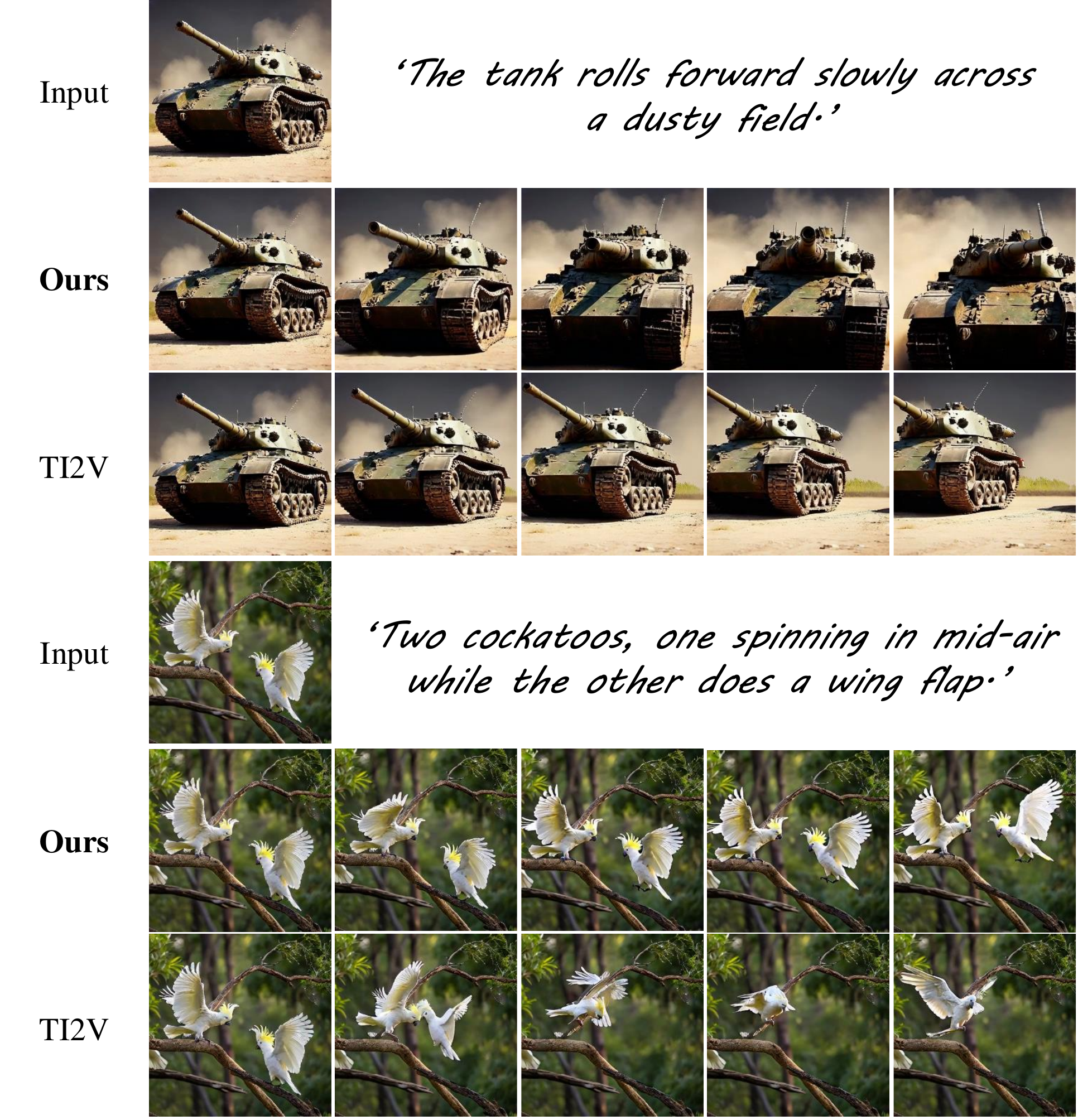}
    \caption{Qualitative comparison of video generations produced by \method (DiT-based) and the TI2V baseline (DiT-based).}
    \label{fig:comp_dit}
\end{figure*}

Building upon the comparisons presented in Sec.~\ref{sec:comparison}, we provide further qualitative results comparing our approach to existing baselines, based on DiT architecture, in Fig.~\ref{fig:comp_dit}

\subsection{Additional Qualitative Comparisons of U-Net Architecture} 
\begin{figure*}[]
  \centering
    \includegraphics[width=\textwidth]{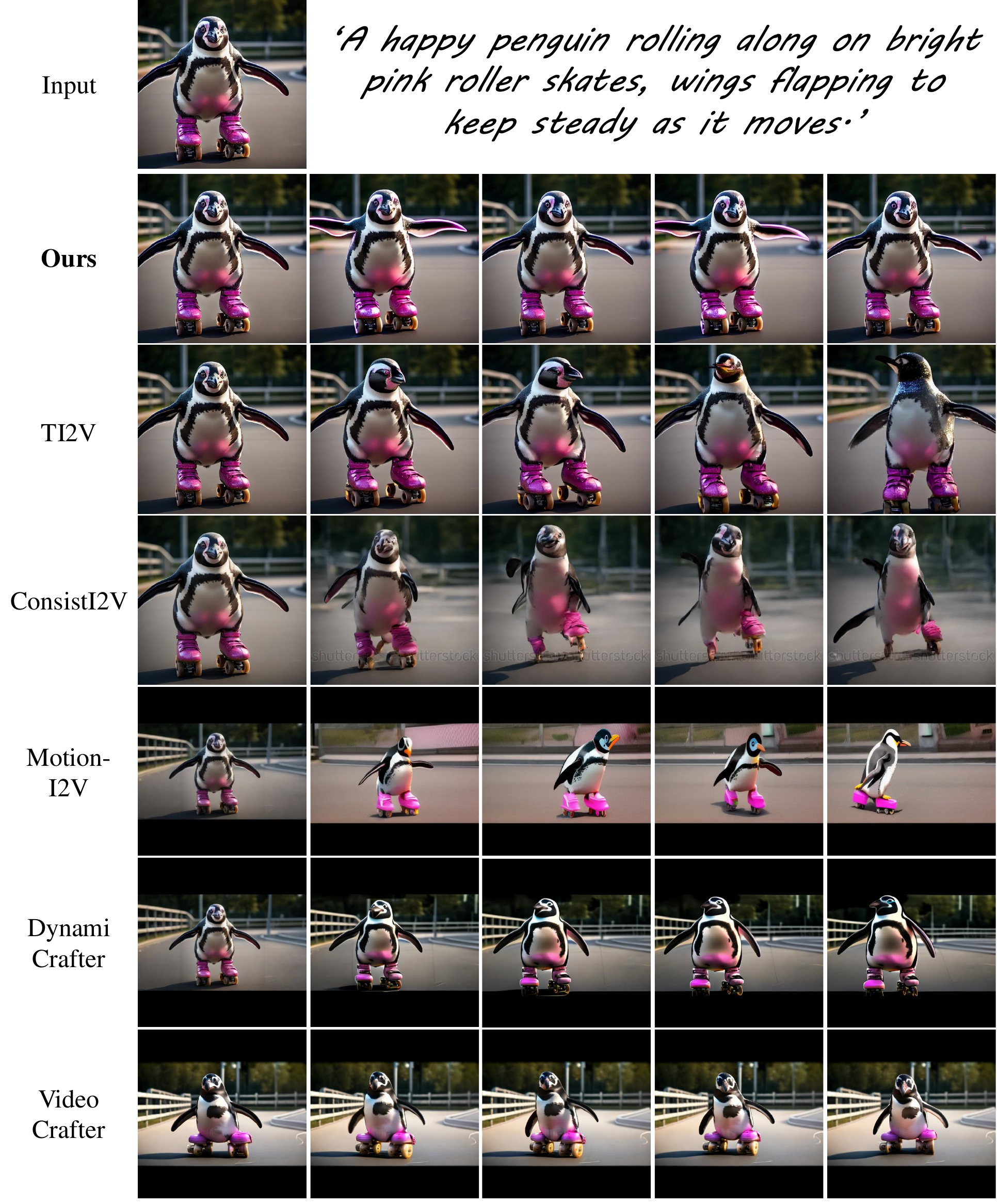}
    \caption{Qualitative comparison of video generations produced by \method (U-Net-based) and TI2V (U-Net-based), ConsistI2V, Motion-I2V, DynamiCrafter, and VideoCrafter.}
    \label{fig:comp_unet}
\end{figure*}

Building upon the comparisons presented in Sec.~\ref{sec:comparison}, we provide further qualitative results comparing our approach to existing baselines, based on U-Net architecture, in Fig.~\ref{fig:comp_unet}

\section{Motion and Object-Specific Prompts Details} 
\label{sec:motion_and_object_prompt}
As described in Sec.~\ref{sec:motion_trajectories}, our pre-processing pipeline extracts a motion-specific prompt, $c_{motion}$, from the input text \( c \), using a pre-trained LLM. This prompt provides a consolidated description of all motion in the scene, excluding any spatial, color, or object-specific details, and serves as a high-level guide for motion generation.

To generate the motion-specific prompt, we use Llama v3.1-8B~\cite{dubey2024llama3herdmodels} in a frozen configuration. The input prompt instructs the LLM to focus solely on motion, as shown in Fig.~\ref{fig:motion_prompt}, ensuring that descriptions remain centered on movement dynamics, ignoring background information and visual characteristics of objects.

\section{Motion-capable Objects' Prompt Extraction Details} 
\label{sec:object_capable}
As described in Sec.~\ref{sec:motion_trajectories}, the pre-processing process begins with extracting motion-capable object prompts from the global prompt \( c \). We utilize Llama v3.1-8B~\cite{dubey2024llama3herdmodels} as a frozen LLM and provide the prompt shown in Fig.~\ref{fig:object_prompt}, which outlines the process for generating local prompts for motion-capable objects.

\section{Inference} 
\label{sec:inference}
Given the reference image $x^{(0)}$ and text prompt $c$, inference is carried out in two stages. 
First, the initial segmentation $s^{(0)}$ is extracted from $x^{(0)}$ using SAM2~\citep{ravi2024sam2segmentimages}.
Concurrently, the text prompt $c$ is processed by a pre-trained LLM to obtain the motion-specific prompt $c_{motion}$ and object-specific prompts $c_{local} = \{c_{local}^{(1)}, \dots, c_{local}^{(L)}\}$ as detailed in Section~\ref{sec:motion_trajectories}.
At stage 1, the image-to-motion generates motion trajectories $\hat{s}$ conditioned on $(s^{(0)}, x^{(0)}, c_{motion})$ .
Next, in stage 2, the motion-to-video produces the final video $\hat{x}$ by conditioning on $(x^{(0)}, \hat{s}, c, c_{local})$ and incorporating masked attention mechanisms to ensure consistency and controllability, as described in Section~\ref{sec:motion_to_video}. 
For both stages, we adapt the Classifier-Free Guidance~\citep{ho2022classifier} approach suggested by~\citet{brooks2023instructpix2pix}, where, to align exactly with their method, we treat the concatenated visual conditions as a single visual condition, and do the same for the text.
% ~\ap{update to match the method symbols}

\section{Implementation Details}
\label{sec:implementation}
As detailed above, we demonstrate the applicability of our approach to two architectures.

The first is the U-Net architecture. We follow the AnimateDiff~V3~\cite{guo2024animatediffanimatepersonalizedtexttoimage} design, consisting of approximately 1.4B parameters. In the second stage of motion-to-video, detailed in Sec.~\ref{sec:motion_to_video}, we set \( K = 6 \), where \( K \) represents the number of attention blocks expanded into masked attention blocks—specifically, by adding masked self-attention and masked cross-attention into the spatial attention blocks within the U-Net's encoder layers. The U-Net-based model was optimized using the solver suggested by~\cite{lin2024commondiffusionnoiseschedules}, incorporating the DDIM diffusion solver with \( v \)-prediction and zero signal-to-noise ratio (SNR). The latter was found to be critically important to enable image-to-mask-based motion trajectory generation.

The second architecture is DiT-based. We train a DiT model following the MovieGen~\cite{polyak2024moviegencastmedia} design, containing four billion parameters. For the DiT-based model in stage two, we used \( K = 10 \), corresponding to the first 10 attention blocks out of a total of 40. The DiT-based model was optimized as described in the MovieGen paper, with Flow Matching~\cite{lipman2023flowmatchinggenerativemodeling}, using a first-order Euler ODE solver. During inference, we adopted MovieGen's efficient inference method by combining a linear-quadratic \( t \)-schedule, as detailed in the MovieGen paper.

For both architectures, text-to-video pre-training followed the methodology outlined in MovieGen. Across both training stages (Sec.~\ref{sec:image_to_motion} and Sec.~\ref{sec:motion_to_video}), we utilized the fine-grained mask-based motion trajectories dataset described in Sec.~\ref{sec:motion_trajectories}. The U-Net model was trained at a resolution of \( 512 \times 512 \), predicting 16 frames, while the DiT model was trained at a resolution of \( 256 \times 256 \), predicting 128 frames. Both models were trained with a batch size of 32, using a constant learning rate of \( 2 \times 10^{-5} \), a warm-up period of 2000 steps, and a total of 50,000 steps.

\section{\benchmark Benchmark} 
\label{sec:benchmark_construction}
We introduce a balanced test set of 128 videos from the SA-V dataset~\cite{ravi2024sam2segmentimages}, comprising 64 single-object and 64 multi-object cases, with an average duration of 14 seconds per video. The filtering of 128 videos, out of the full SA-V dataset, involved several steps. First, for each video, we generated a text caption using Llama v3.2-11B~\cite{dubey2024llama3herdmodels} by providing the first, middle, and last frames and asking the model to generate a caption describing the video. Next, from a closed set of categories (Animal, Architecture, Digital Art, Food, Landscape, Lifestyle, Plant, Vehicles, Visual Art, and Other), we used Llama v3.2-11B~\citep{dubey2024llama3herdmodels} to categorize each video based on these frames. We then iterated over the categories, selecting a unique category at each step and adding a related video to ensure a balanced test set. We assigned an aesthetic score and a motion score by calculating the magnitude of the optical flow extracted with RAFT~\cite{teed2020raftrecurrentallpairsfield}. After assigning captions and scores, we filtered 500 videos by iterating through each category and selecting those with the highest combined aesthetic and motion scores. From these 500 automatically filtered videos, we randomly selected 64 single-object and 64 multi-object videos. To ensure a fair comparison for shorter video settings, we also provided short captions, generated using the same methodology, extracted from frames 0 to 127 of each video.

\section{Image-Animation-Bench} 
\label{sec:i2v_bench}
The Image-Animation-Bench comprises 2,500 videos, meticulously curated to meet high-resolution requirements and aesthetic quality thresholds. To ensure comprehensive coverage of diverse visual scenarios, the dataset is divided into 16 categories: Portraits, Scenery-Nature, Pets, Food, Animation, Science, Sports, Scenery-City, Animation-Static, Music, Game, Animals, Industry, Painting, Vehicles, and others.

\begin{figure*}[]
\begin{verbatim}
Task: Extract a single motion-specific prompt from the caption that describes 
the overall motion without including any spatial, color, size, or background details.

Format your answer like this:
Motion-specific prompt: "description of overall motion"

Examples:

Caption: "A large, red ball rolls to the right on a grassy field while a small, 
blue kite flies upward in the clear, blue sky."
Motion-specific prompt: "The ball rolls to the right, and the kite flies upward."

Caption: "A sleek, black car drives down a busy city street with tall buildings 
in the background as several pedestrians wearing bright clothing cross."
Motion-specific prompt: "The car drives down the street as pedestrians cross."

Caption: "A fluffy, white cat jumps onto a wooden table set against a plain, 
beige wall and knocks over a glass of water, spilling it onto the floor."
Motion-specific prompt: "The cat jumps onto the table and knocks over the glass."

Now, please provide the answer.
Caption: "{global_prompt}"
Motion-specific prompt: 
\end{verbatim}
\caption{The input prompt used for extracting a motion-specific description from the global prompt \( c \), designed for use with a pre-trained LLM. The prompt focuses solely on describing the overall motion, explicitly excluding details such as sizes, colors, or background elements. Here, \( c \) refers to the global prompt, which is inserted in place of \texttt{\{global\_prompt\}}.}
\label{fig:motion_prompt}
\end{figure*}

\begin{figure*}[]
\begin{verbatim}
Task: For each object mentioned in the caption, write a local prompt that describes 
everything about that object as mentioned in the caption.

Format your answer like this:
Answer: [[Object 1: description of object 1] [Object 2: description of object 2] ...]

Examples:

Caption: "An alien rides a horse through a field."
Answer: [[alien: A alien rides a horse through a field.] 
         [horse: A horse is being ridden through a field.]]

Caption: "A dog chases a ball while a robot runs after it."
Answer: [[dog: A dog chases a ball.] 
         [ball: A ball is being chased by a dog.] 
         [child: A robot runs after it.]]

Caption: "An eagle flies above the mountains."
Answer: [[eagle: The eagle flies above the mountains.]]

Caption: "Two playful dogs run along the beach, with one dog on the left and the other
          in the middle of the frame, as waves crash onto the shore."
Answer: [[left dog: The dog runs playfully along the beach, staying closer to the dry sand.] 
         [middle dog: The dog runs beside its companion, edging nearer to the waves.]]

Caption: "Three cats sit in a row on a sunny windowsill, all basking in the warm sunlight,
          when the cat on the right starts to move his paw."
Answer: [[left cat: The cat sits on the windowsill, soaking in the sunlight.] 
         [middle cat: The cat sits on the windowsill, soaking in the sunlight.] 
         [right cat: The cat sits on the windowsill, then starts to move his paw.]]

Caption: "A bustling farmers' market filled with a variety of colorful fruit stands, 
where a monkey is carefully picking ripe, red tomatoes while a street musician 
plays lively tunes on an acoustic guitar, adding a vibrant atmosphere to the scene."
Answer: [[monkey: A monkey carefully picks ripe, red tomatoes from one of the stands.] 
         [musician: A street musician plays lively tunes on an acoustic guitar]]
Now, please provide the answer.
Caption: "{global_prompt}"
Answer: 
\end{verbatim}
\caption{The input prompt used for extracting motion-capable object descriptions from the global prompt \( c \), designed for use with a pre-trained LLM. Here, \( c \) refers to the global prompt, which is inserted in place of \texttt{\{global\_prompt\}}.}
\label{fig:object_prompt}
\end{figure*}

\end{document}